\documentclass[10pt,twocolumn,letterpaper]{article}

\usepackage{cvpr}              % To produce the CAMERA-READY version
\makeatletter
\@namedef{ver@everyshi.sty}{}
\makeatother
\usepackage{tikz}

\usepackage{graphicx}
\usepackage{amsmath}
\usepackage{amssymb}
\usepackage{booktabs}
\usepackage{amsfonts}
\usepackage{wrapfig}
\usepackage{times}
\usepackage{multirow}
\usepackage{url}
\usepackage{color}
\usepackage{xcolor}
\usepackage{soul}

\usepackage{adjustbox}
\usepackage{lipsum}
\usepackage{color}
\usepackage{wrapfig}
\usepackage{booktabs}
\usepackage{multirow,mathtools} 
\usepackage{algorithm,algpseudocode}
\usepackage{threeparttable}
\usepackage{amsthm}

\usepackage{enumitem}
\usepackage{xspace}
\usepackage{xfrac}
\usepackage{comment}
\usepackage{tikz}

\usepackage{etoolbox}
\newtoggle{mydebug}
\settoggle{mydebug}{true}

\usepackage[skip=0pt, figurename=Fig.]{caption}
\usepackage[skip=0pt, tablename=Tab.]{caption}

\usepackage[pagebackref,breaklinks,colorlinks]{hyperref}

\usepackage[capitalize]{cleveref}
\crefname{section}{Sec.}{Secs.}
\Crefname{section}{Section}{Sections}
\Crefname{table}{Table}{Tables}
\crefname{table}{Tab.}{Tabs.}

 \newcommand{\ours}{\text{ILM-VP}}

\DeclareMathOperator*{\argmax}{arg\,max}

\DeclareMathOperator*{\minimize}{minimize}
\usepackage[most]{tcolorbox}

\usepackage{pifont}
\usepackage{adjustbox}
\usepackage{threeparttable,booktabs}
\usepackage{multirow}

\DeclareMathAlphabet\mathbfcal{OMS}{cmsy}{b}{n}

\newcommand{\bdelta}{\boldsymbol{\delta}}

\DeclareMathOperator*{\ST}{\text{subject to}}

\usepackage{amsmath}%
\usepackage{MnSymbol}%
\usepackage{wasysym}%

\usepackage{tcolorbox}

\usepackage{mathtools}
\DeclarePairedDelimiterX{\inp}[2]{\langle}{\rangle}{#1, #2}

\usepackage{color, colortbl}
\definecolor{Gray}{gray}{0.93}
\definecolor{Orange}{rgb}{1,0.5,0}
\definecolor{DGray}{gray}{0.83}
\definecolor{LightCyan}{rgb}{0.88,1,1}

\usepackage{{tcolorbox}}
\newcommand{\mycomment}[1]{}
\usepackage{xcolor,colortbl,pifont}
\newcommand*\colourcheck[1]{%
  \expandafter\newcommand\csname #1check\endcsname{\textcolor{#1}{\ding{52}}}%
}
\colourcheck{blue}
\colourcheck{green}
\colourcheck{red}
\newcommand*\colourcross[1]{%
  \expandafter\newcommand\csname #1cross\endcsname{\textcolor{#1}{\ding{56}}}%
}
\colourcross{red}

 \newcommand{\bthetasrc}{\boldsymbol \theta_{\mathrm{s}}}
 \newcommand{\Ysrc}{\mathcal Y_{\mathrm{s}}}
  \newcommand{\Ytgt}{\mathcal Y_{\mathrm{t}}}

 \newcommand{\xtgt}{\mathbf x_{\mathrm{t}}}
  \newcommand{\ytgt}{y_{\mathrm{t}}}
 
  \newcommand{\ysrc}{y_{\mathrm{s}}}
 \newcommand{\Nsrc}{N_{\mathrm{s}}}
 \newcommand{\Ksrc}{K_{\mathrm{s}}}
 \newcommand{\Ntgt}{N_{\mathrm{t}}}
 \newcommand{\Ktgt}{K_{\mathrm{t}}}

\def\cvprPaperID{620} % *** Enter the CVPR Paper ID here
\def\confName{CVPR}
\def\confYear{2023}

\usepackage{color}

\begin{document}

\title{Understanding and Improving Visual Prompting: A Label-Mapping Perspective}

\author{Aochuan Chen$^1$, ~~Yuguang Yao$^1$, ~~Pin-Yu Chen$^2$, ~~Yihua Zhang$^1$,~~ Sijia Liu$^{1,2}$\\
$^1$Michigan State University, ~~$^2$MIT-IBM Watson AI Lab, IBM Research\\
%{\tt\small chenaoch@msu.edu}
}

\maketitle

%%%%%%%%% ABSTRACT
\begin{abstract}
We revisit and advance visual prompting (\textbf{VP}), an input prompting technique for vision tasks. VP can \textbf{reprogram} a fixed, pre-trained source model to accomplish downstream tasks in the target domain by simply incorporating universal prompts (in terms of input perturbation patterns) into downstream data points. Yet, it remains elusive why VP stays effective even given a \textbf{ruleless} label mapping (\textbf{LM}) between the source classes and the target classes.  Inspired by the above, we ask: 
How is LM interrelated with VP? And how to exploit such a relationship to improve its accuracy on target tasks?
We peer into the influence of LM on VP and provide an affirmative answer that a better `quality' of LM (assessed by mapping precision and explanation) can consistently improve the effectiveness of VP. This is in contrast to the prior art where the factor of LM was missing. {To optimize LM,}
we propose a new VP framework, termed {\textbf{\ours}} (\underline{i}terative \underline{l}abel \underline{m}apping-based \underline{v}isual \underline{p}rompting), which automatically re-maps the source labels to the target labels 
and progressively improves the target task accuracy of VP.
Further, when using a contrastive language–image pretrained (CLIP) model for VP, we propose to integrate an LM process to assist the text prompt selection of CLIP and to improve the target task accuracy.
Extensive experiments demonstrate that our proposal significantly outperforms state-of-the-art VP methods. As highlighted below,  we show that when reprogramming an ImageNet-pretrained ResNet-18 to 13 target tasks, {\ours} outperforms baselines by a substantial margin, \textit{e.g.}, 7.9\% and 6.7\% accuracy improvements in transfer learning to the target Flowers102 and CIFAR100 datasets. Besides, our proposal on CLIP-based VP provides 13.7\% and 7.1\% accuracy improvements on Flowers102 and DTD respectively. 
Code is available at https://github.com/OPTML-Group/ILM-VP.
\end{abstract}

%%%%%%%%% BODY TEXT
\vspace{-7mm}
\section{Introduction}
\label{sec: intro}

\vspace{-1.5mm}
\begin{figure}
    \centering
    \includegraphics[width=\columnwidth]{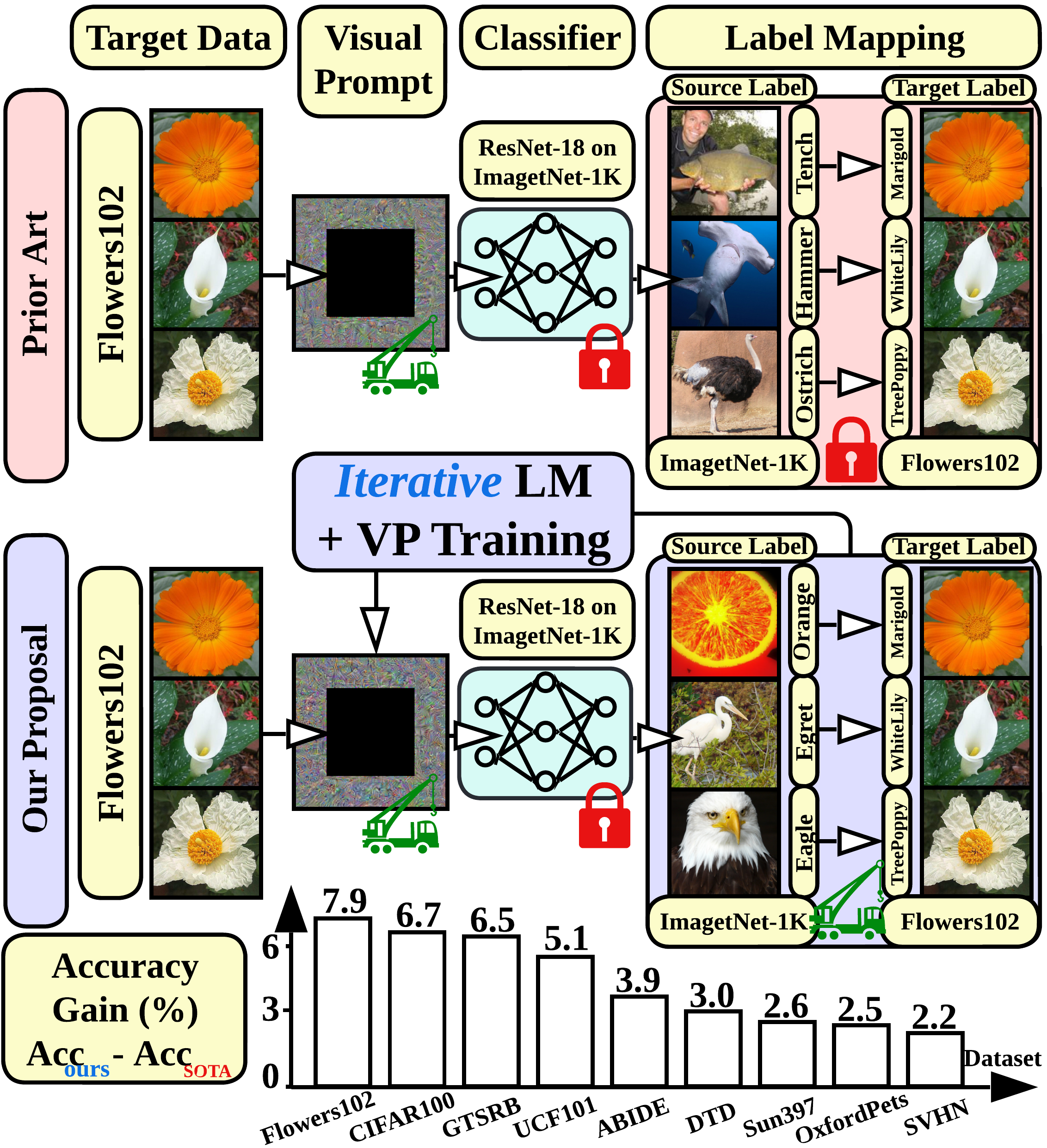}
    \caption{
    \footnotesize{
   Overview of VP pipelines (prior art \cite{tsai2020transfer,bahng2022visual} and our proposal termed {\ours}) and accuracy improvement achieved by {\ours} on target image classification tasks at-a-glance. Generally speaking,   VP   aims to generate a universal input perturbation template (\textit{i.e.}, `visual prompt') and leverage a source-target LM (label mapping) in order to drive the fixed source model (\textit{e.g.}, pretrained on ImageNet-1K) to conduct a target task (\textit{e.g.}, Flowers102 image classification).  Compared to the prior art, our proposal ({\ours}) couples the design of LM with VP training. The resulting LM-VP co-design improves target task accuracy across a variety of target image classification tasks using a fixed ImageNet-pretrained source model.
    }
    }
    \label{fig: ILM_overview}
    \vspace*{-6mm}
\end{figure}

%Transfer learning is a well-known technique in machine learning (ML).

When learning new knowledge,   humans typically start to compare and connect it with the knowledge that they were familiar with. 
The same idea is also applied in ML. For example, in the `pretraining + finetuning' paradigm,  an ML model (\textit{e.g.}, deep neural network or DNN) is first trained on a (usually large) \textit{source} dataset. When a relevant downstream task is present, the pre-trained model is then fine-tuned over the \textit{target} dataset.  This learning paradigm has been predominant in the classical transfer learning  \cite{pan2010domain, pan2009survey, ghifary2014domain, tzeng2014deep, daume2009frustratingly, maitra2015cnn} as well as in the recent deep representation learning \cite{dai2008self, raina2007self, chen2020simple, chen2021exploring, khosla2020supervised}.

However, finetuning the pre-trained model requires either partial or entire model modifications. If the pre-trained model is of large size, then it becomes too costly to store a modified copy of the pre-trained model for each downstream task. In contrast, visual prompting (\textbf{VP}) (see \textbf{Fig.\,\ref{fig: ILM_overview}}), also known as model reprogramming or adversarial reprogramming,  provides a  new alternative to finetuning \cite{bahng2022visual, elsayed2018adversarial, tsai2020transfer, chen2021adversarial, neekhara2018adversarial, neekhara2022cross}.   Instead of directly modifying the pre-trained source model, VP integrates an \textit{input transformation} and/or an \textit{output transformation}  to reprogram the \textit{fixed} source model to accomplish a new target task; see an illustration of existing VP framework in \textbf{Fig.\,\ref{fig: ILM_overview}}. The input transformation is typically realized by incorporating (data-agnostic) input perturbations (\textit{i.e.},  prompts) into input samples, and the output transformation is given by a function that maps source labels to target labels, known as label mapping  (\textbf{LM}).
Recently,  VP has shown great promise in various applications of foundation models, ranging from pre-trained vision models \cite{tsai2020transfer, elsayed2018adversarial, neekhara2022cross, chen2021adversarial, bar2022visual, sohn2022visual, gao2022visual} to language-vision models \cite{bahng2022visual, xing2022class, zang2022unified, khattak2022maple}.

The idea of prompt learning originated from in-context learning or prompting in natural language processing (NLP)  \cite{brown2020language,li2021prefix,radford2021learning}. However, when it is introduced to the vision domain \cite{tsai2020transfer,bahng2022visual}, new questions arise. \textbf{First}, the recent work \cite{elsayed2018adversarial,tsai2020transfer,yen2021study}  showed that VP remains powerful even if the target task largely deviates from the source domain. For example,  a new performance record on target medical datasets is achieved in \cite{tsai2020transfer} when using VP to reprogram the fixed,  ImageNet pre-trained source model. The `mystery' in this example is that LM is conducted between two seemingly irrelevant source and target domains.
Despite the lack of interpretability, VP can still leverage such connected source labels and the source model to effectively predict target data points. 
This raises the first open question: \textit{What is the rationality behind LM  and how to explore its influence on VP?} 
\textbf{Second}, unlike prompt learning in the NLP domain, input prompts in the vision domain are typically given by `noisy' perturbations to image pixels; see illustration in \textbf{Fig.\,\ref{fig: ILM_overview}}. Together with the lack of interpretability of LM, the second open question is: \textit{How to interpret LM and the seemingly random perturbation pattern in VP?}

As mentioned above, 
the lack of understanding of LM and the poor interpretability of VP  drive our studies in this work. We develop a new visual prompting framework, termed \textbf{\ours} (\underline{i}terative \underline{l}abel \underline{m}apping-based \underline{v}isual \underline{p}rompting), which provides an interactive and explainable design between LM and prompt learning (\textit{i.e.}, input prompt generation); see \textbf{Fig.\,\ref{fig: ILM_overview}} for the schematic overview. Our proposal can automatically adjust LM between the source domain and the target domain by taking both mapping precision and explanation into consideration, and can leverage the optimized LM to further improve the accuracy and the explainability of prompt learning.
Although some {prior work}  \cite{tsai2020transfer, neekhara2022cross, yen2021study} attempted to improve the quality of LM as well as the overall performance of VP, they are different from our proposal in two major aspects. \textbf{First}, none of the prior work co-designed LM and VP.  For example, the prior art
\cite{tsai2020transfer} used a pre-prompt prediction frequency to determine the LM function. However, we find significant inconsistency between the pre-prompt and post-prompt prediction frequency of the same source model, which explains the sub-optimality of the current  VP methods due to the lack of mapping precision.  \textbf{Second}, to the best of our knowledge, 
VP is still treated as a `black box' in the prior work. Yet, our design can provide graceful visual explanations to the underlying mechanisms of VP. \textbf{Third}, we for the first time show that  LM can provide a unified solution to improving the accuracy of VP to re-purpose both vision and language-vision source models.
Our \textbf{contributions} are unfolded below.

\ding{172}  We revisit the LM problem in VP and uncover the deficiencies of existing LM methods: the lack of mapping precision and the lack of explanation. 

\ding{173} Given the importance of LM, we propose the first LM-VP co-design framework, termed {\ours}, through a novel bi-level optimization viewpoint. 

\ding{174} Beyond LM for vision models, we show that LM can also be generalized to assist the text prompt selection of CLIP (contrastive language–image pretraining) and to improve the target task accuracy of VP using the CLIP model.

\ding{175} We empirically demonstrate the accuracy and explanation merits of our proposal across multiple source models and target datasets.  
\vspace{-3mm}
\section{Related Work}
\label{sec: literature}

\vspace{-2mm}
\paragraph{Prompting in  NLP.} Prompting is used to prepend language instruction to the input text for a language model to better accomplish a given task\cite{liu2021pre}. While prompting makes a significant contribution to the generalization ability of large pre-trained language models (\textit{e.g.}, GPT-3) \cite{brown2020language}, it requires hand-crafting prompt design by experts.
Recent work proposed to directly optimize the prompting embeddings through gradients together with lightweight finetuning the model, which is called \textit{prompt tuning}\cite{li2021prefix, lester2021power}. It is shown that this method is effective and efficient, which achieves competitive performance to the finetuning of the full language model. 

\vspace{-6mm}
\paragraph{Visual prompting and model reprogramming.}
VP was first defined in \cite{bahng2022visual} to mimic the prompting idea in NLP. 
Prior to that, a very similar idea was used in computer vision (CV) but with a different name, known as \textit{model reprogramming or adversarial reprogramming} \cite{elsayed2018adversarial, chen2022model, zheng2021adversarial, neekhara2018adversarial, neekhara2022cross, chen2021adversarial,zhang2022fairness,chen2022visual}. They both focus on re-purposing a fixed, pre-trained vision model for a new task by leveraging a universal input pattern and an output LM function. Although not outperforming full fine-tuning in transfer learning, VP yields an advantage of parameter-efficient fine-tuning, which requires a much smaller parameter storage space. Furthermore, the smaller parameter space requires less training data to converge.
Beyond traditional pre-trained vision models, the work \cite{bahng2022visual} studied the effectiveness of VP in the language-vision model CLIP for the first time. Assisted by CLIP, VP can generate a prompting pattern of image data without resorting to source-target label mapping. 
In \cite{khattak2022maple}, VP and {text prompt}   are jointly optimized in the CLIP model, which leads to better performance. Furthermore,
\textit{unadversarial learning} \cite{salman2021unadversarial} also enjoys the similar idea to VP, while it focuses on generating class-wise prompts with the goal of improving the out-of-distribution generalization ability of a pre-trained model.

VP is gaining increasing attention. 
In \cite{tsai2020transfer}, 
it is applied to re-purpose black-box source models\cite{zhang2022robustify} and achieves state-of-the-art (SOTA) performance on different target datasets. Besides, in data-scarce regimes like the biochemical domain, 
it is shown in \cite{yen2021study, chen2021adversarial, neekhara2022cross} that VP can enable effective cross-domain transfer learning. Other than transfer learning, VP is also used in in-domain settings to improve different metrics like adversarial robustness \cite{chen2022visual} and fairness \cite{zhang2022fairness}.
Although input prompting is the most commonly-used prompt learning method in the vision domain, generalization to learning prompting parameters at intermediate layers of a source model is also developed in \cite{jia2022visual,xing2022class, gao2022visual, sohn2022visual}. The resulting technique is called visual prompt tuning and is typically restricted to vision transformers.

\vspace{-4mm}
\section{Problem Statement}
\label{sec: motivation}
\vspace{-2mm}
In this section, we begin by providing some background information on VP.
Based on that, we will then present the problem of our interest--LM (label mapping)--which defines how a visual prompt maps a
source model prediction label to a target data class. This is the first question encountered in VP across domains but was typically overlooked in the literature. By reviewing the commonly-used LM methods, we will point out several open questions raised by LM.

\vspace{-6mm}
\paragraph{Preliminaries on visual prompting.}
The technology of VP   addresses the problem of how to adapt a pre-trained \textit{source} model (\textit{e.g.}, the ImageNet-1K-pre-trained ResNet-18) to a \textit{target} downstream task (\textit{e.g.}, flower classification over the Flowers102 dataset) \textit{without} any task-specific model modification (\textit{e.g.}, finetuning).
Throughout the paper, we focus on input-based VP  (also known as {model reprogramming}) \cite{tsai2020transfer, bahng2022visual, elsayed2018adversarial, chen2021adversarial, neekhara2022cross, neekhara2018adversarial, yen2021study, zhang2023text}, which incorporates a   carefully-designed universal perturbation pattern to the raw target images so as to enforce the transferability of the source model to the target domain. We refer readers to \textbf{Fig.\,\ref{fig: ILM_overview}} for the schematic overview. 

To be concrete, let $\mathcal S$ and $\mathcal T$ denote the source dataset and the target dataset, respectively. And let $f_{\bthetasrc} $ denote a \textit{{s}ou{rc}e model} with pre-trained parameters $\bthetasrc$. Suppose  $f_{\bthetasrc} $ is a \textit{supervised classifier}, then it defines a mapping from the input data $\mathbf x \in \mathbb R^{\Nsrc} $ to the source label space $\Ysrc   \subseteq \mathbb R^{\Ksrc}$, \textit{i.e.}, $f_{\bthetasrc}( \mathbf x ) = \ysrc \in \Ysrc $, where $\Nsrc$ is the dimension of a source datapoint, $\Ksrc$ is the number of source data classes, and $\ysrc$ is the source class label. We have $f_{\bthetasrc} $   trained based on $\mathcal S$, \textit{e.g.}, via empirical risk minimization. 
\textbf{The goal of VP} is   to re-program  the source model $f_{\bthetasrc} $ to accomplish the target task defined in  $\mathcal T$, without making task-specific finetuning over $f_{\bthetasrc} $. To this end, VP   modifies the target data $\xtgt $ (of $\Ntgt$ dimensions) by injecting a   task-designated input perturbation pattern $\boldsymbol \delta$. This leads to the \textbf{input prompting operation} with the generic form:

{
\vspace*{-4.5mm}
\small
{
\begin{align}
    \mathbf x^\prime (\boldsymbol \delta) = h ( \xtgt, \boldsymbol \delta  ) \in \mathbb R^{\Nsrc},  ~
    \xtgt  \in   \mathbb R^{\Ntgt}
    \label{eq: input_prompting}
\end{align}
}}%
where $\xtgt$ is the target datapoint, and $h(\cdot, \cdot)$ is an input transformation that integrates $\xtgt$ with the input perturbation $\boldsymbol \delta$ and produces a modified datapoint $\mathbf x^\prime (\boldsymbol \delta)$ with the source data dimension $\Nsrc$. It was shown in \cite{tsai2020transfer} that $h$ can be specified as an additive perturbation model that pads $\boldsymbol \delta$ outside the target data sample (see \textbf{Fig.\,\ref{fig: ILM_overview}} for an example as well). 

Given the input prompting model \eqref{eq: input_prompting}, VP then seeks the optimal $\boldsymbol \delta$ to improve the target task accuracy when using the pre-trained source model  $f_{\bthetasrc}$.
This raises a \textbf{prompt generation problem}, which is typically cast as

{
\vspace*{-4.5mm}
\small
{
\begin{align}
    \begin{array}{ll}
   \displaystyle \minimize_{\boldsymbol \delta}      & \mathbb E_{(\xtgt, \ytgt ) \in \mathcal T_{\mathrm{tr}}} \, [ \ell_\mathrm{VP}( f_{\bthetasrc} (  \mathbf x^\prime %_{\mathrm{t}} 
   (\boldsymbol \delta)) , \ytgt  ) ]  ,
    \end{array}
    \label{eq: prob_prompt_generation}
\end{align}
}}%
where $\mathcal T_{\mathrm{tr}}$ denotes a supervised training set in $\mathcal T$ with  feature $\xtgt$ and label $\ytgt$ for a training sample, and  $\ell_\mathrm{VP}(\cdot)$ is a  visual prompting loss function that we will define later given the prompted input $\mathbf x^\prime %_{\mathrm{t}} 
   (\boldsymbol \delta)$ and the ground-truth target label $\ytgt$.
   To solve problem \eqref{eq: prob_prompt_generation}, the standard stochastic gradient descent (SGD) method can be used.  At inference, we will integrate the designed  $\bdelta$    into test-time target datapoints and call the source model $f_{\bthetasrc}$ for downstream prediction in $\mathcal{T}$ (see \textbf{Fig.\,\ref{fig: ILM_overview}} and a more detailed description in \textbf{Fig.\,\ref{fig: VP_overview}}).

\vspace{-4.5mm}
\paragraph{Label mapping: Existing methods and questions.}

Although the input prompting operation \eqref{eq: input_prompting} converts the original $\xtgt$ to the source dimension-aligned datapoint $\mathbf x^\prime$ that the source model can use, the successful realization of VP \eqref{eq: freq_LM}   needs to map the source model's prediction (in the source label space $\Ysrc$ with $\Ksrc$ classes) to the target task's data label (in the target label space $\Ytgt$ with $\Ktgt$ classes). In the `pretraining + finetuning' paradigm,
 we typically have $\Ktgt \leq \Ksrc$. %\textit{e.g.},   the setup of ImageNet (source dataset) + GTSRB (target dataset) as shown in Fig.\,\SL{[RLMFLMBipartite]}.
Therefore, the problem of LM (label mapping) arises:
\begin{tcolorbox}
\vspace*{-2mm}
(\textbf{LM problem})  Given the source model $f_{\bthetasrc}$, how to build a mapping from the source label space  $\Ysrc$ to the target label space $\Ytgt$  so that the model's prediction directs to the correct target label?
\vspace*{-2.5mm}
\end{tcolorbox}

Clearly, the desired prompt generation     \eqref{eq: prob_prompt_generation} heavily relies on the LM scheme, which defines the one-to-one correspondence between the source model's prediction $f_{\bthetasrc} (  \mathbf x^\prime (\boldsymbol \delta) ) $ and the target data class $\ytgt$.
Yet, nearly all the existing work neglects its influence on the prompt generation and adopts either \ding{172} the simplest random mapping \cite{elsayed2018adversarial,bahng2022visual} or \ding{173} a pre-defined, one-shot frequency-based mapping \cite{tsai2020transfer, chen2021adversarial}. We elaborate on the above two schemes below. 

\noindent \ding{172} \textbf{Random label mapping (RLM)}: RLM does  \textit{not} use any prior knowledge or source model information to guide the LM process. The mapped source labels (to the target domain) could be even random. For example, in the case of `ImageNet (source) + CIFAR-10 (target)'  \cite{elsayed2018adversarial,bahng2022visual}, existing VP  methods coded CIFAR-10 labels using the top 10 ImageNet indices, \textit{i.e.}, ImageNet label $i$ $\to$ CIFAR-10 label $i$, despite the lack of interpretation. 

\noindent \ding{173} \textbf{Frequency-based label mapping (FLM)}:  FLM matches target labels to source labels based on the source model's prediction frequencies on zero-padded target datapoints, \textit{i.e.}, $f_{\bthetasrc} ( \mathbf x^\prime (\bdelta))$ with $\bdelta = \mathbf 0$. Here recall from \eqref{eq: input_prompting} that  $\mathbf x^\prime (\mathbf 0) = h ( \xtgt, \mathbf 0  )
$. More concretely,
 a target label $\ytgt$ is mapped to the  source label $\ysrc^*$ following

{
\vspace*{-4.5mm}
\small
{
\begin{align}
\hspace*{-5mm}  \begin{array}{l}
      \ysrc^* (\ytgt) = \argmax_{\ysrc}\, \mathrm{Pr}  \left \{   \text{Top-1 prediction of } f_{\bthetasrc} ( h ( \xtgt, \mathbf 0  ) )   
      \right.  \\
     \left.  \hspace*{32mm} 
     \text{ is } 
    {\ysrc} 
    \, | \,  \forall \xtgt \in \mathcal T_{\ytgt} \right  \} 
\end{array}
  \hspace*{-5mm}  \label{eq: freq_LM}
\end{align}
}}%
where $\ysrc^* (\ytgt)$ explicitly expresses the dependence of the mapped source label on the target label,   
$\mathcal T_{\ytgt}$ denotes the  target data set in the class  $\ytgt$, and $\mathrm{Pr}\{ \cdot \} $ is the probability of the event that the top-1 prediction of $ f_{\bthetasrc} $ is  the source class $\ysrc$ under the zero-padded target data points in  $\mathcal T_{\ytgt}$.

\begin{figure}
    \centering
    \includegraphics[width=\columnwidth]{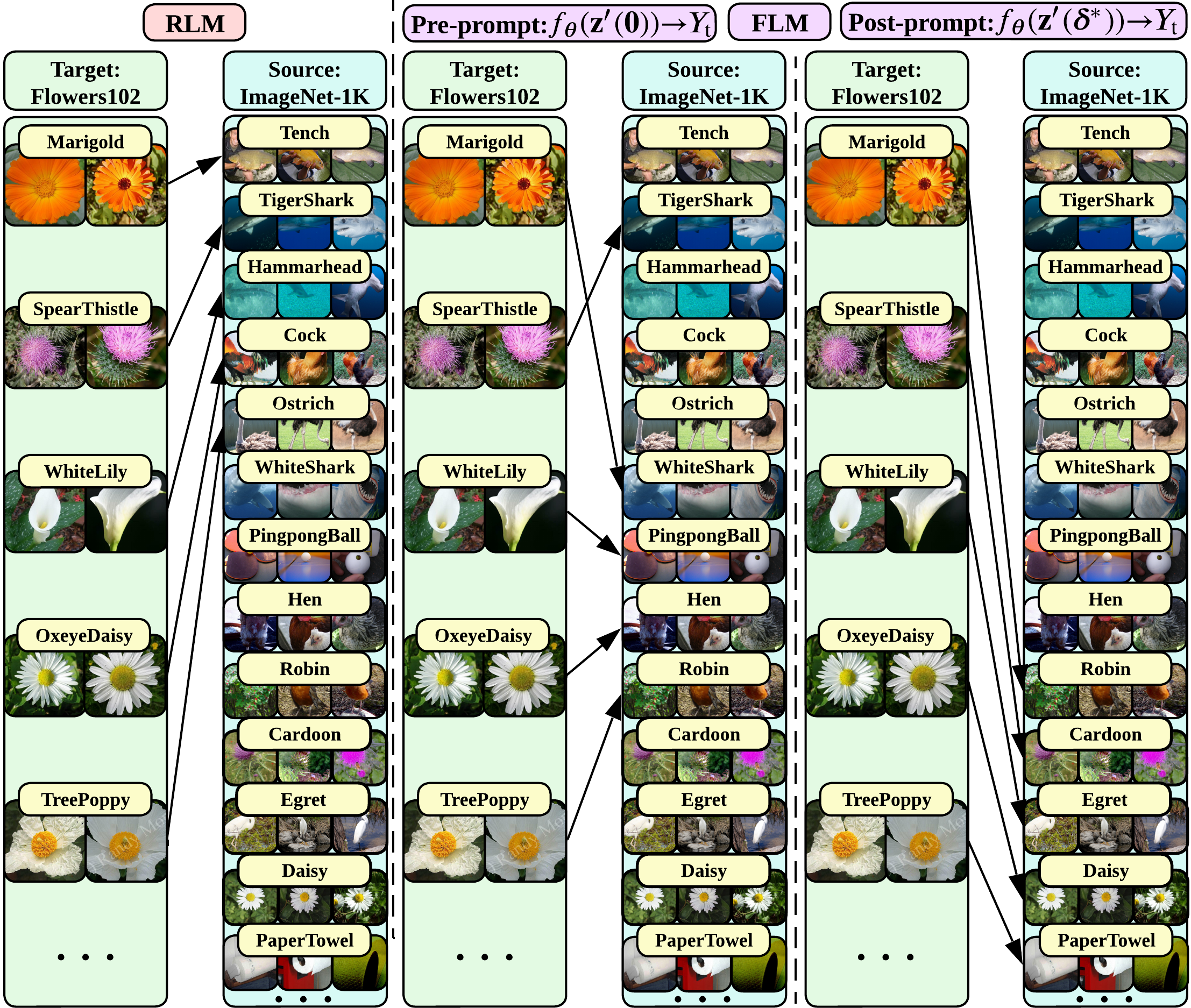}
    \caption{\footnotesize{Visualizations of RLM and FLM   using the source dataset ImageNet-1K and the pretrained ResNet-18, as well as 
    the target dataset Flowers102. In FLM, 
    %Although FLM exhibits more interpretability than RLM, 
    the pre-prompt label mapping using \eqref{eq: freq_LM} selects source labels different from FLM. Yet, the post-prompt label mapping using \eqref{eq: freq_LM} at $\boldsymbol \delta^*$ in \eqref{eq: dynamics_source_label} shows
    {many} newly-selected source labels, indicating (1) the dynamics of LM in the source domain, and (2) the pre-promp LM is sub-optimal  (\textit{i.e.}, mis-selecting the best-matching label) after VP training.}}
    \label{fig: bipartite_RLM_FLM}
    \vspace{-7mm}
\end{figure}

As shown in \textbf{Fig.\,\ref{fig: bipartite_RLM_FLM}},  FLM results in a   mapping scheme different from that of RLM. However, it is still difficult to interpret the obtained LM results, and remains elusive how the quality of LM impacts the performance of VP. 
In the rest of the work, we will shed light on how to improve VP by carefully designing the LM scheme and when and why LM matters at different source and downstream tasks.
\vspace{-4mm}
\section{Method: Iterative Label Mapping-based VP}
\label{sec: method}
\vspace{-2mm}
In this section, we uncover the \textit{hidden dynamics of LM} existing in the source task domain of VP, which was neglected by the prior art.  
This finding then motivates us to develop a novel VP framework, which we
call {I}terative {LM}-based {VP} ({\textbf{\ours}}).
Compared to existing VP methods, {\ours} closes the loop between LM and prompt generation \eqref{eq: prob_prompt_generation}, and improves  VP's explanation and target task accuracy simultaneously. 

\vspace{-4.5mm}
\paragraph{The `missing' dynamics of LM in the source domain.}
As shown in Sec.\,\ref{sec: motivation},   a prompt learning pipeline   mainly involves three steps: \textbf{(A1)} input prompt modeling \eqref{eq: input_prompting}, \textbf{(A2)} LM (from the source label set $\Ysrc$ to the target label set $\Ytgt$), and  \textbf{(A3)} prompt generation \eqref{eq: prob_prompt_generation}.
The prior art follows the  pipeline {(A1)}$\to${(A2)}$\to${(A3)} to generate the desired prompt   $\boldsymbol \delta^*$, which drives the source model to accomplish {target}   tasks.
However, in the viewpoint of the \textit{source} domain, the prompt updating from   $\boldsymbol \delta = \mathbf 0$ to $\boldsymbol \delta^*$   induces the {prediction dynamics} of the source model $ f_{\bthetasrc}$. That is,

{\small{
\vspace*{-6mm}
\begin{align}
 f_{\bthetasrc} (  \mathbf x^\prime (\mathbf 0 )   ) \to f_{\bthetasrc} (
  \mathbf x^\prime (\boldsymbol \delta^*) 
 ),
   \label{eq: dynamics_source_label}
\end{align} }}%
where $ \mathbf x^\prime (\boldsymbol \delta )$ has been defined in  \eqref{eq: input_prompting}, which refers to the $\boldsymbol \delta$-perturbed target data
with the same dimension as the source datapoint.
As will be evident later, it is important to understand the dynamics \eqref{eq: dynamics_source_label} as it  reflects the stability of the selected source labels
when mapping to the target labels.

\textbf{Fig.\,\ref{fig: bipartite_RLM_FLM}} instantiates the dynamics of  \eqref{eq: dynamics_source_label} in the scenario of `ImageNet (source) + Flowers102 (target)’  when   the FLM-oriented VP approach \eqref{eq: freq_LM} is used \cite{tsai2020transfer}.
Prior to prompt generation, 
the Flowers102 target labels are first mapped to the ImageNet source labels using 
the FLM method  \eqref{eq: freq_LM}, corresponding to step (A2) in prompt learning. This yields the \textit{pre-prompt} target-source mapping, denoted by 
$f_{\bthetasrc} (
 \mathbf x^\prime (\mathbf 0) ) \stackrel{\mathclap{\normalfont\mbox{\tiny FLM}}}{\to}  \Ytgt$. 
 Similarly, after generating the prompt   $\boldsymbol \delta^*$   following (A3), we can  obtain  the \textit{post-prompt} target-source mapping, $f_{\bthetasrc} (
 \mathbf x^\prime (\boldsymbol \delta^*) ) \stackrel{\mathclap{\normalfont\mbox{\tiny FLM}}}{\to}  \Ytgt$, using the FLM method. \textbf{Fig.\,\ref{fig: bipartite_RLM_FLM}}  shows that there exists a significant    \textit{discrepancy} between the pre-prompt LM and the post-prompt LM, evidenced by the newly-selected source labels (`Cardoon', `Egret', `Daisy', `Paper Towel') in the post-prompting phase. This justifies the dynamics of \eqref{eq: dynamics_source_label} in LM. 
   However, it also raises a \textit{new concern}   that 
 the pre-prompt target-source LM is \textit{sub-optimal} for prompt generation \eqref{eq: prob_prompt_generation} given the existing dynamics of LM in the source domain.

\vspace{-6mm}
\paragraph{ILM-VP: A bi-level optimization viewpoint of VP.}
The dynamics of LM inspire us to re-think the optimality of the current VP pipeline: 
{(A1)}$\to${(A2)}$\to${(A3)}. To improve it,  \textbf{we propose}   to take the LM dynamics into the prompt learning process. This modifies the conventional VP pipeline to {(A1)}$\to${(A2)}$\rightleftarrows${(A3)}, where LM and prompt generation are in a closed loop. 
Since the design of LM will interact with the design of the prompt iteratively, we call the proposed new design \textbf{{\ours}}.

Next, we formally present  ILM-VP through the lens of bi-level optimization (\textbf{BLO}).  Generally speaking, BLO provides a hierarchical learning
framework involving two levels (\textit{i.e.}, upper and lower levels) of optimization tasks, where one task is nested inside the other (\textit{i.e.}, the objective and variables of an upper-level problem depend on the optimizer of
the lower-level problem). In the context of ILM-VP,  
we regard the prompt generation problem \eqref{eq: prob_prompt_generation} as 
the upper-level optimization task and the LM problem  \eqref{eq: freq_LM} as the lower-level problem. This yields

{
\vspace*{-4.5mm}
\small
{
 \begin{align}
 % \displaystyle
   \hspace*{-3mm}  \begin{array}{l}
\underbrace{\displaystyle \minimize_{\boldsymbol \delta}       \,\,\, \mathbb E_{(\xtgt, \ytgt ) \in \mathcal T_{\mathrm{tr}}} \, [ \ell( f_{\bthetasrc} (  \mathbf x^\prime  (\boldsymbol \delta)), \ysrc^*(\ytgt)  ) ] }_\text{Upper-level prompt optimization}  \vspace*{0.5mm}\\
  \ST     \,\,\, \displaystyle \underbrace{\ysrc^*(\ytgt) \text{ is obtained by \eqref{eq: freq_LM}  at   (non-zero) prompt $\bdelta$}
 }_\text{Lower-level LM design at current prompt $\boldsymbol \delta$ for every target label $\ytgt$}
    \end{array}
    \hspace*{-3mm}
    \label{eq: BLO_ILM}
\end{align}
}}%
where the visual prompt $\boldsymbol \delta$  denotes the upper-level variable,  $\ell$ is the cross-entropy loss,  and the mapped source label $\ysrc$ is a lower-level variable for each given target label $\ytgt$ at the current prompt $\bdelta$. We also note that there exists a
lower-level constraint in \eqref{eq: BLO_ILM} to ensure that
if a source class has been mapped to a target class, it will then be excluded when mapping to a new target class. 
Further, it is clear from \eqref{eq: BLO_ILM} that the design of visual prompt $\bdelta$ and LM $\ysrc^*$ (vs. $\ytgt$) are intertwined with each other. 
   
To solve problem \eqref{eq: BLO_ILM}, we employ the alternating optimization (AO) method, which alternatively executes the upper-level prompt generation and the lower-level LM. We summarize the algorithm details in Algorithm\,\ref{alg: bi_ilm} and provide a schematic overview in \textbf{Fig.\,\ref{fig: algorithm_overview}}.% \SL{[use $n$ to represent epoch number]}

\vspace{-3.5mm}
\begin{algorithm}[H]
\caption{The proposed {\ours} algorithm}
\begin{algorithmic}[1]
  \State \textbf{Initialize:} Given target training set $\mathcal{T}_\mathrm{tr}$, pre-trained model $f_{{\bthetasrc}}$, prompt pattern initialization $\boldsymbol{\delta}_0$, and upper-level learning rate $\lambda$ for SGD
  \For{Epoch $n=0,1,\ldots,$}
\State \textbf{Lower-level label mapping}: Given $\boldsymbol{\delta}_{n-1}$,  call   LM
    \hspace*{5.5mm}for  each 
target class $\ytgt$ in $\mathcal{T}_\mathrm{tr}$
\State \textbf{Upper-level prompt learning}: Given LM, call 
\hspace*{5.5mm}SGD to update prompt
$\bdelta_n \xleftarrow{} \bdelta_{n-1}$
\EndFor
\end{algorithmic}
\label{alg: bi_ilm}
\end{algorithm}

\begin{figure}
    \centering
    \includegraphics[width=\columnwidth]{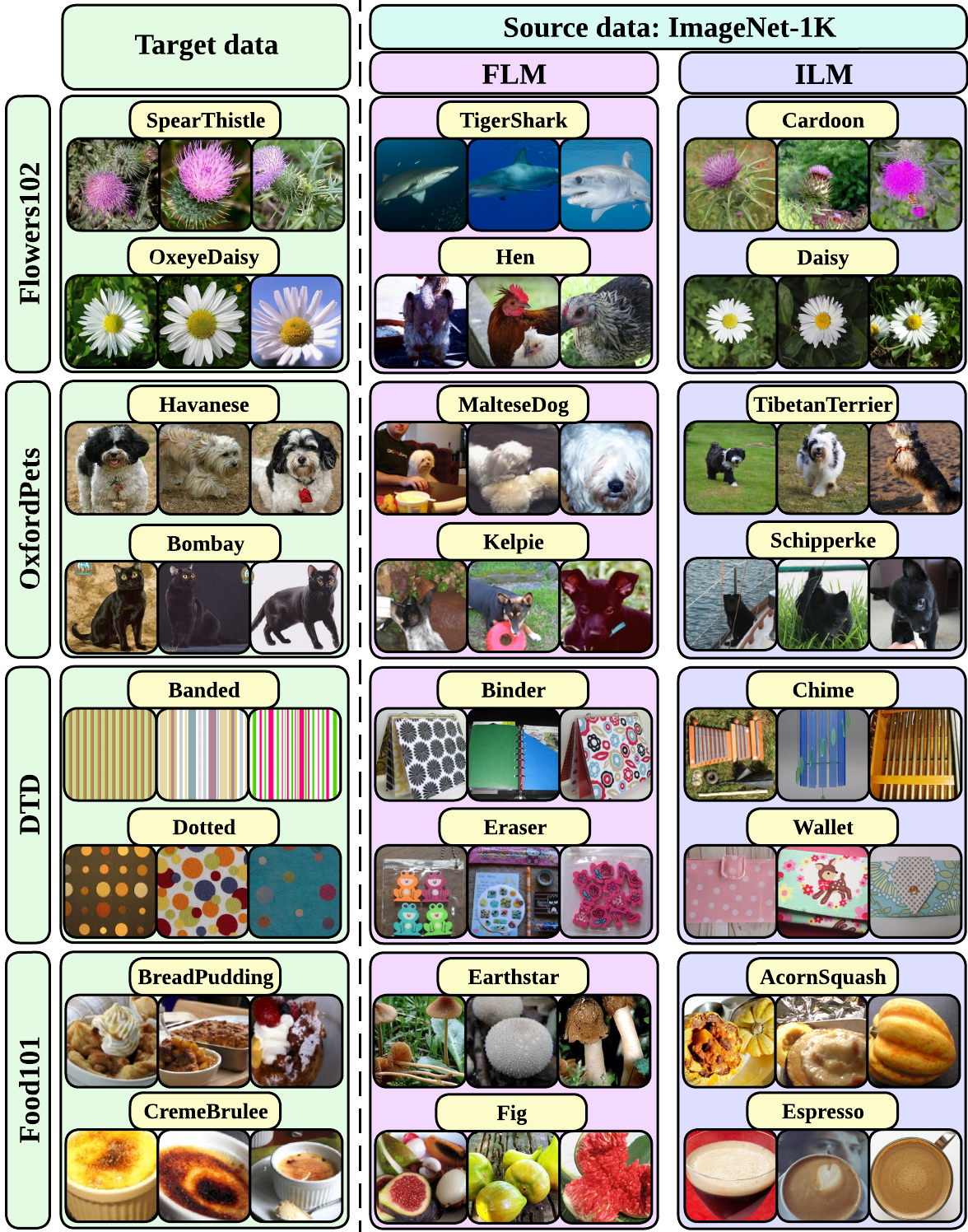}
    \caption{\footnotesize{Interpretation merit of ILM (ours) vs. FLM, visualized by LM results in VP to re-purpose an ImageNet-pretrained source model (ResNet-18) to conduct target image classification tasks on the target datasets Flowers102, OxfordPets, DTD, and Food101. ILM consistently finds more interpretable target-source label mappings than FLM, in terms of colors, scenes, shapes, and textures. See \textbf{Fig.\,\ref{fig: more_FLM_ILM}} for more examples.}}
    \label{fig: visualization_FLM_ILM}
    \vspace*{-7mm}
\end{figure}

\vspace{-10.1mm}
\paragraph{An interpretation merit of ILM-VP.}
In the literature, it is quite difficult to interpret why VP can reprogram a source model to conduct target tasks. 
The main hurdle of interpreting VP  lies in the LM phase: It remains elusive why the semantics-irrelevant source labels should be mapped to target labels. However, we find that ILM-VP can alleviate this interpretation difficulty to a large extent. We show the explanation merit of ILM-VP through an empirical study in \textbf{Fig.\,\ref{fig: visualization_FLM_ILM}}, where
the target dataset is instantiated by Flowers102 and the source dataset is ImageNet-1K. We list the target labels, the mapped source labels using the baseline FLM  method  \cite{tsai2020transfer}, and the identified source labels using ILM-VP, together with image examples under each label.   As we can see,  an \textit{interpretable} target-source mapping is found by ILM-VP, even if the target label and the source label describe different subjects.  For example,  target images in the label `Spear Thistle' share a similar color and object shape with the source images in the label `Cardoon'. The same observations can also be drawn from other target-source label mappings together with their data instances. This finding is quite encouraging and is in sharp contrast to FLM. As will be evident in \textbf{Sec.\,\ref{sec: exp}}, the BLO-oriented ILM-VP  \eqref{eq: BLO_ILM} would enforce a convergence of LM as the alternating optimization proceeds. As a result, source labels and target labels, which share the most similar concepts (like colors, scenes, shapes, and materials), will be identified. We also show that the improved interpretation of LM consistently enhances the target task accuracy of the VP.

\vspace{-4mm}
\section{Experiments}
\label{sec: exp}
\vspace{-2mm}

\begin{table*}[htb]
\centering
%\vspace*{-2em}
\resizebox{\textwidth}{!}{%
\begin{tabular}{c|c|cc|cc|c|c|cc|c|c|cc}
\toprule[1pt]
\midrule
Source Model & \multicolumn{5}{c|}{ResNet-18 (ImageNet-1K)} & \multicolumn{4}{c|}{ResNet-50 (ImageNet-1K)} & \multicolumn{4}{c}{ResNeXt-101-32x8d (Instagram)} \\
\midrule
 \multirow{2}{*}{Method} & \multicolumn{1}{c|}{\underline{Ours}} & \multicolumn{2}{c|}{Prompt baseline} & \multicolumn{2}{c|}{Finetuning} & \multicolumn{1}{c|}{\underline{Ours}} & \multicolumn{1}{c|}{Prompt base.} & \multicolumn{2}{c|}{Finetuning} & \multicolumn{1}{c|}{\underline{Ours}} & \multicolumn{1}{c|}{Prompt base.} & \multicolumn{2}{c}{Finetuning} \\
  & \multicolumn{1}{c|}{\ours} & \multicolumn{1}{c}{RLM-VP} & \multicolumn{1}{c|}{FLM-VP} & \multicolumn{1}{c}{LP} & \multicolumn{1}{c|}{FF} & \multicolumn{1}{c|}{\ours} & \multicolumn{1}{c|}{FLM-VP} & \multicolumn{1}{c}{LP} & \multicolumn{1}{c|}{FF} & \multicolumn{1}{c|}{\ours} & \multicolumn{1}{c|}{FLM-VP} & \multicolumn{1}{c}{LP} & \multicolumn{1}{c}{FF}\\
\midrule
Parameter Size & \multicolumn{1}{c|}{0.05M} & \multicolumn{1}{c}{0.05M} & \multicolumn{1}{c|}{0.05M} & \multicolumn{1}{c}{0.51M} & \multicolumn{1}{c|}{11.7M} & \multicolumn{1}{c|}{0.05M} & \multicolumn{1}{c|}{0.05M} & \multicolumn{1}{c}{0.51M} & \multicolumn{1}{c|}{25.6M} & \multicolumn{1}{c|}{0.05M} & \multicolumn{1}{c|}{0.05M} & \multicolumn{1}{c}{0.51M} & \multicolumn{1}{c}{88.8M} \\
\midrule
Flowers102
& \textbf{27.9\footnotesize{$\pm0.7$}}
& 11.0\footnotesize{$\pm0.5$}  
& 20.0\footnotesize{$\pm0.3$} 
& 88.0\footnotesize{$\pm0.5$} 
& 97.1\footnotesize{$\pm0.7$} 
& \textbf{24.6\footnotesize{$\pm0.6$}}
& 20.3\footnotesize{$\pm0.3$} 
& 90.9\footnotesize{$\pm0.4$} 
& 97.9\footnotesize{$\pm0.7$} 
& \textbf{27.9\footnotesize{$\pm0.3$}}
& 22.5\footnotesize{$\pm0.5$} 
& 89.1\footnotesize{$\pm0.2$} 
& 99.2\footnotesize{$\pm0.5$} 
\\
DTD
& \textbf{35.3\footnotesize{$\pm0.9$}}
& 16.3\footnotesize{$\pm0.7$}  
& 32.4\footnotesize{$\pm0.5$} 
& 60.0\footnotesize{$\pm0.6$} 
& 65.5\footnotesize{$\pm0.9$} 
& \textbf{40.5\footnotesize{$\pm0.5$}} 
& 36.9\footnotesize{$\pm0.8$} 
& 67.6\footnotesize{$\pm0.3$} 
& 69.7\footnotesize{$\pm0.9$} 
& \textbf{41.4\footnotesize{$\pm0.7$}}
& 40.3\footnotesize{$\pm0.5$} 
& 69.7\footnotesize{$\pm0.2$} 
& 69.1\footnotesize{$\pm1.0$} 
\\
UCF101
& \textbf{23.9\footnotesize{$\pm0.5$}}
& 6.6\footnotesize{$\pm0.4$}  
& 18.9\footnotesize{$\pm0.5$} 
& 63.2\footnotesize{$\pm0.8$} 
& 73.0\footnotesize{$\pm0.6$}
& \textbf{34.6\footnotesize{$\pm0.2$}} 
& 33.9\footnotesize{$\pm0.4$} 
& 70.8\footnotesize{$\pm0.3$} 
& 78.0\footnotesize{$\pm0.8$} 
& \textbf{43.1\footnotesize{$\pm0.8$}} 
& 41.9\footnotesize{$\pm0.6$} 
& 76.9\footnotesize{$\pm0.5$} 
& 79.1\footnotesize{$\pm0.7$} 
\\
Food101
& \textbf{14.8\footnotesize{$\pm0.2$}}
& 3.8\footnotesize{$\pm0.3$}  
& 12.8\footnotesize{$\pm0.1$} 
& 50.6\footnotesize{$\pm0.3$} 
& 75.4\footnotesize{$\pm0.8$} 
& \textbf{17.0\footnotesize{$\pm0.3$}}
& 15.3\footnotesize{$\pm0.2$} 
& 57.6\footnotesize{$\pm0.5$} 
& 80.3\footnotesize{$\pm0.9$} 
& \textbf{23.0\footnotesize{$\pm0.4$}} 
& 20.5\footnotesize{$\pm0.5$} 
& 76.0\footnotesize{$\pm0.4$} 
& 82.5\footnotesize{$\pm0.3$} 
\\
GTSRB
& \textbf{52.0\footnotesize{$\pm1.2$}}
& 46.1\footnotesize{$\pm1.3$}  
& 45.5\footnotesize{$\pm1.0$} 
& 77.4\footnotesize{$\pm1.2$} 
& 98.0\footnotesize{$\pm0.3$} 
& \textbf{52.5\footnotesize{$\pm1.4$}} 
& 47.6\footnotesize{$\pm1.1$} 
& 77.8\footnotesize{$\pm0.7$} 
& 97.6\footnotesize{$\pm1.0$} 
& \textbf{59.9\footnotesize{$\pm1.0$}} 
& 56.2\footnotesize{$\pm0.6$} 
& 73.5\footnotesize{$\pm0.7$} 
& 97.6\footnotesize{$\pm0.9$} 
\\
EuroSAT
& \textbf{85.2\footnotesize{$\pm0.6$}}
& 82.4\footnotesize{$\pm0.4$}  
& 83.8\footnotesize{$\pm0.2$} 
& 93.8\footnotesize{$\pm0.3$} 
& 98.8\footnotesize{$\pm0.5$} 
& 83.6\footnotesize{$\pm0.7$} 
& \textbf{84.8\footnotesize{$\pm0.3$}} 
& 95.7\footnotesize{$\pm0.2$} 
& 98.9\footnotesize{$\pm0.6$} 
& 86.2\footnotesize{$\pm0.8$} 
& \textbf{87.8\footnotesize{$\pm0.4$}} 
& 93.4\footnotesize{$\pm0.3$} 
& 98.9\footnotesize{$\pm0.7$} 
\\
OxfordPets
& \textbf{65.4\footnotesize{$\pm0.7$}}
& 9.3\footnotesize{$\pm0.4$}  
& 62.9\footnotesize{$\pm0.1$} 
& 87.2\footnotesize{$\pm0.6$} 
& 87.8\footnotesize{$\pm0.5$} 
& 76.2\footnotesize{$\pm0.6$} 
& \textbf{76.4\footnotesize{$\pm0.2$}} 
& 90.4\footnotesize{$\pm0.3$} 
& 91.9\footnotesize{$\pm0.4$} 
& \textbf{78.9\footnotesize{$\pm0.8$}} 
& 76.8\footnotesize{$\pm0.6$} 
& 93.6\footnotesize{$\pm0.4$} 
& 90.1\footnotesize{$\pm0.9$} 
\\
StanfordCars
& \textbf{4.5\footnotesize{$\pm0.1$}}
& 0.9\footnotesize{$\pm0.1$}  
& 2.7\footnotesize{$\pm0.1$} 
& 33.8\footnotesize{$\pm0.2$}
& 81.0\footnotesize{$\pm0.1$} 
& \textbf{4.7\footnotesize{$\pm0.2$}} 
& 4.2\footnotesize{$\pm0.3$} 
& 40.6\footnotesize{$\pm0.1$} 
& 86.4\footnotesize{$\pm0.3$} 
& \textbf{7.0\footnotesize{$\pm0.2$}} 
& 4.6\footnotesize{$\pm0.1$} 
& 64.7\footnotesize{$\pm0.1$} 
& 92.5\footnotesize{$\pm0.2$} 
\\
SUN397
& \textbf{13.0\footnotesize{$\pm0.2$}}
& 1.0\footnotesize{$\pm0.1$}  
& 10.4\footnotesize{$\pm0.1$} 
& 46.1\footnotesize{$\pm0.2$} 
& 53.2\footnotesize{$\pm0.2$} 
& \textbf{20.3\footnotesize{$\pm0.2$}} 
& 19.8\footnotesize{$\pm0.1$} 
& 53.5\footnotesize{$\pm0.1$} 
& 59.0\footnotesize{$\pm0.1$} 
& \textbf{23.7\footnotesize{$\pm0.2$}} 
& 21.6\footnotesize{$\pm0.3$} 
& 62.3\footnotesize{$\pm0.1$} 
& 61.0\footnotesize{$\pm0.2$} 
\\
CIFAR10
& 65.5\footnotesize{$\pm0.1$}
& 63.0\footnotesize{$\pm0.1$}  
& \textbf{65.7\footnotesize{$\pm0.6$}}
& 85.9\footnotesize{$\pm0.5$} 
& 96.5\footnotesize{$\pm0.4$} 
& \textbf{76.6\footnotesize{$\pm0.3$}} 
& 74.8\footnotesize{$\pm0.5$} 
& 90.1\footnotesize{$\pm0.1$} 
& 96.6\footnotesize{$\pm0.2$} 
& \textbf{81.7\footnotesize{$\pm0.3$}} 
& 80.3\footnotesize{$\pm0.3$} 
& 94.1\footnotesize{$\pm0.1$} 
& 97.1\footnotesize{$\pm0.1$} 
\\
CIFAR100
& \textbf{24.8\footnotesize{$\pm0.1$}}
& 12.9\footnotesize{$\pm0.1$}  
& 18.1\footnotesize{$\pm0.2$} 
& 63.3\footnotesize{$\pm0.8$} 
& 82.5\footnotesize{$\pm1.2$} 
& \textbf{38.9\footnotesize{$\pm0.3$}} 
& 32.0\footnotesize{$\pm0.4$} 
& 70.7\footnotesize{$\pm0.7$} 
& 83.4\footnotesize{$\pm0.9$} 
& \textbf{45.9\footnotesize{$\pm0.2$}} 
& 39.7\footnotesize{$\pm0.2$} 
& 76.2\footnotesize{$\pm0.9$} 
& 84.6\footnotesize{$\pm1.2$} 
\\
SVHN
& \textbf{\color{black} 75.2\footnotesize{$\pm0.2$}}
& 73.5\footnotesize{$\pm0.3$}  
& 73.1\footnotesize{$\pm0.2$} 
& 65.0\footnotesize{$\pm0.2$} 
& 96.5\footnotesize{$\pm0.3$} 
& \textbf{\color{black} 75.8\footnotesize{$\pm0.4$}} 
& 75.6\footnotesize{$\pm0.2$} 
& 63.5\footnotesize{$\pm0.2$} 
& 96.9\footnotesize{$\pm0.3$} 
& \textbf{81.4\footnotesize{$\pm0.1$}} 
& 79.0\footnotesize{$\pm0.5$} 
& 51.0\footnotesize{$\pm0.2$} 
& 97.1\footnotesize{$\pm0.3$} 
\\
ABIDE
& \textbf{\color{black} 76.9\footnotesize{$\pm2.1$}}
& 74.0\footnotesize{$\pm2.2$}  
& 73.1\footnotesize{$\pm1.6$} 
& 65.4\footnotesize{$\pm3.8$} 
& 60.6\footnotesize{$\pm4.2$} 
& 63.5\footnotesize{$\pm2.2$} 
& \textbf{64.4\footnotesize{$\pm3.4$}} 
& 55.8\footnotesize{$\pm2.6$} 
& 70.2\footnotesize{$\pm2.5$} 
& \textbf{67.3\footnotesize{$\pm2.6$}} 
& 65.7\footnotesize{$\pm3.4$} 
& 54.8\footnotesize{$\pm3.4$} 
& 73.1\footnotesize{$\pm4.2$} 
\\
\midrule
\bottomrule[1pt]
\end{tabular}%
}
\caption{\footnotesize{Performance overview of our proposed VP method ({\ours}), prompt baseline methods (RLM-VP and FLM-VP), and finetuning methods (LP and FF) over $13$ target image classification datasets using $3$ pretrained source models (ResNet-18 on ImageNet-1K, ResNet-50 on ImageNet-1K, and ResNeXt-101-32x8d on Instagram).  In each cell, a\footnotesize{$\pm b$} refers to the mean and standard deviation of target task accuracies (\%) over 3 independent trials. The highest accuracy across VP-based methods is marked in \textbf{bold}.
 `Parameter Size' refers to the number of trainable parameters in the input prompt or model finetuning.
}}
\label{tab: exp_main}
\vspace*{-7mm}
\end{table*}

In this section, we empirically demonstrate the effectiveness of our proposed {\ours} method by comparing it with a variety of baselines across multiple datasets, models, and learning paradigms. 

\vspace{-3mm}
\subsection{Experiment setups}
\vspace{-1.5mm}
\paragraph{Datasets and models.}
In the source domain, we will consider the source models ResNet-18 and ResNet-50 \cite{he2016deep} pre-trained on ImageNet-1K \cite{deng2009imagenet}, and the source model ResNeXt-101-32x8d \cite{xie2017aggregated} pre-trained on Instagram\cite{mahajan2018exploring}. 
In the target domain, 
we  will evaluate the  performance of   {\ours} over \textbf{13 target datasets}: Flowers102\cite{nilsback2008automated}, DTD\cite{cimpoi2014describing}, UCF101\cite{soomro2012ucf101}, Food101\cite{bossard2014food}, GTSRB\cite{Houben-IJCNN-2013}, SVHN\cite{netzer2011reading}, EuroSAT\cite{helber2019eurosat}, OxfordPets\cite{parkhi2012cats}, StanfordCars\cite{KrauseStarkDengFei-Fei_3DRR2013}, SUN397\cite{xiao2010sun}, CIFAR10/100\cite{krizhevsky2009learning}, ABIDE\cite{craddock2013neuro}.

\begin{figure}[t]
    \centering
     \begin{subfigure}[b]{0.49\columnwidth}
         \centering
         \includegraphics[width=\textwidth]{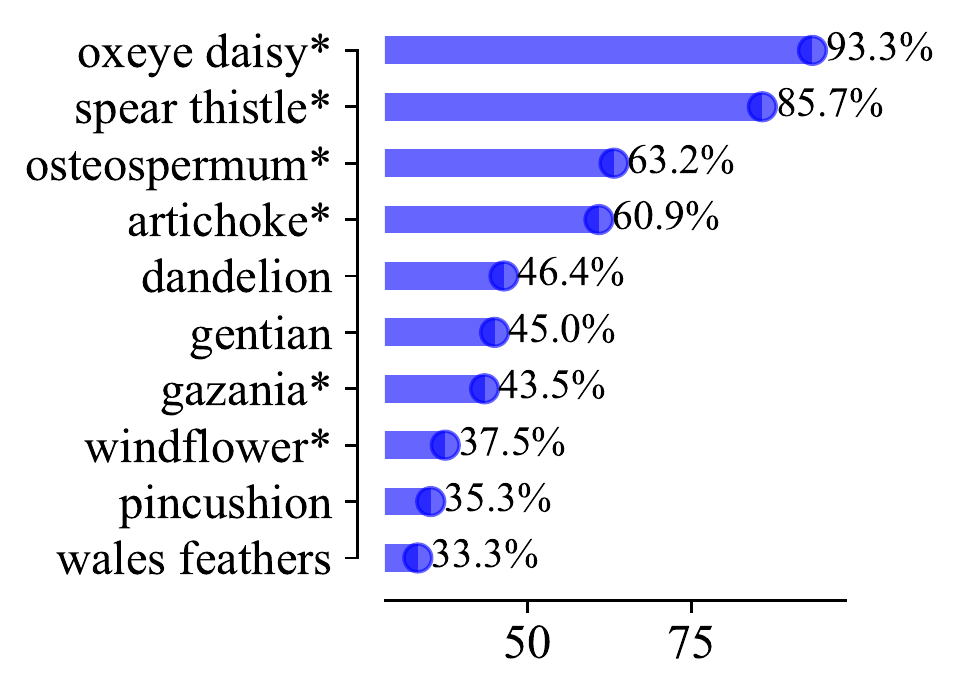}
         \vspace*{-5mm}
         \caption{\footnotesize{Flowers102}}
         \label{fig: cls_acc_flowers}
     \end{subfigure}
     \begin{subfigure}[b]{0.49\columnwidth}
         \centering
         \includegraphics[width=\textwidth]{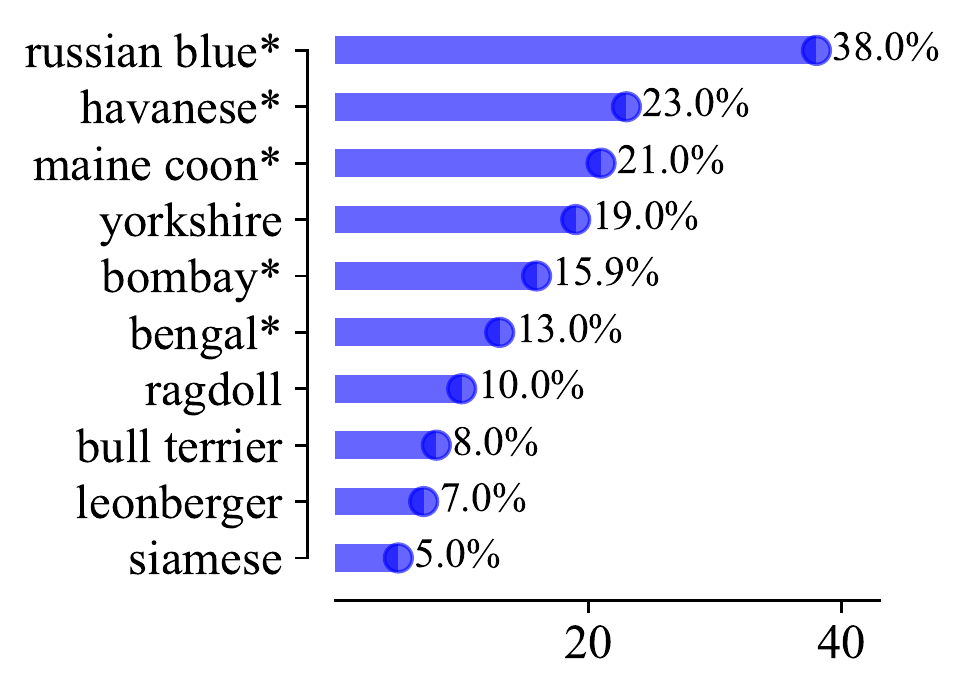}
         \vspace*{-5mm}
         \caption{\footnotesize{OxfordPets}}
         \label{fig: cls_acc_oxford_pets}
     \end{subfigure}
     \iffalse
     \begin{subfigure}[b]{0.33\textwidth}
         \centering
         \includegraphics[width=\textwidth]{AChen/figs/dtd_cls_acc_sample.pdf}
         \caption{\footnotesize{DTD}}
         \label{fig: cls_acc_dtd}
     \end{subfigure}
     \fi
     \caption{\footnotesize{{\ours}'s class-wise accuracy improvements over FLM-VP  
     using ImageNet-1K pre-trained ResNet-18 under the target dataset (a) Flowers102, (b) OxfordPets. Target classes with $*$ refer to those with updated source labels compared to the prompting baseline method FLM-VP.}}
     \label{fig: cls_acc}
     \vspace*{-8mm}
\end{figure}

\vspace{-4.5mm}
\paragraph{Baselines and evaluations.}
In the VP paradigm, our baseline methods include the random LM-based VP (\textbf{RLM-VP}) \cite{elsayed2018adversarial, bahng2022visual}, and the frequency LM-based VP (\textbf{FLM-VP})\cite{tsai2020transfer, chen2021adversarial}. We highlight that   VP   is a \textit{finetuning-free} method to drive the source model to conduct target image classification tasks.  When implementing VP baselines, we follow their official repository setups. We also refer readers to Appendix\,\ref{sec: additional_experiments_and_details} for detailed implementations of {\ours} and baseline methods.
In addition to prompting methods, we also cover finetuning-based methods, including linear probing (\textbf{LP}) and end-to-end full finetuning (\textbf{FF}). Since finetuning modifies source model parameters, it requires training more parameters and is more computationally intensive.

We evaluate the   performance
of all the methods by the target task accuracy at the testing time and the efficiency in terms of the parameter size that VP or finetuning needs to handle. 
We also leverage a post-hoc model explanation method, known as  Explanation-by-Example (\textbf{EBE})  \cite{jeyakumar2020can}, to assess the quality of visual explanation of different VP methods. 
The core idea of EBE is to find train-time data points that have the most similar feature representations to that of a queried test datapoint so as to use these identified training samples to explain the model's prediction on this test sample.
In the context of VP, EBE can aid us to find the source training samples explainable for model prediction on prompted target test data, like source examples in \textbf{Fig.\,\ref{fig: visualization_FLM_ILM}}.

\begin{figure*}[t]
    \centering
    \includegraphics[width=1.8\columnwidth]{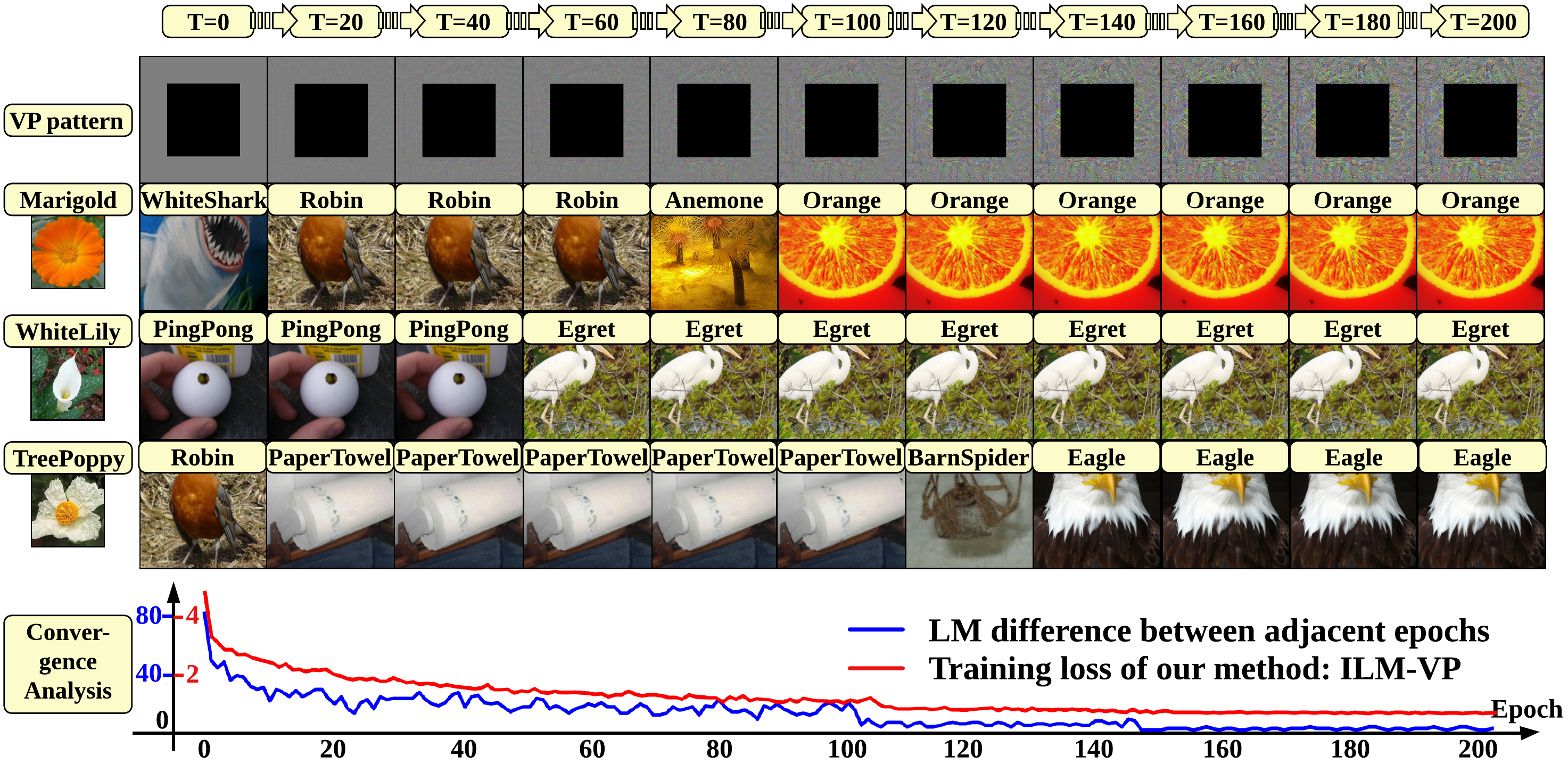}
    \caption{\footnotesize{{\ours} training dynamics from epoch $0$ to $200$. Rows show: (1) VP pattern vs. epoch number; (2-4) Learned source label mapping with respect to target label `Marigold', `White Lily', and `Tree Poppy', together with EBE-identified source training examples to explain each re-purposed target label; (5) Convergence of training loss and LM difference between adjacent epochs measured by Hamming distance.
    }
    }
    \label{fig: ILM_gradient}
    \vspace{-7mm}
\end{figure*}

\vspace{-2.5mm}
\subsection{Experiment results}
\label{sec: experiment_results}
\vspace{-3mm}
\paragraph{Overall performance of {\ours}.}
\textbf{Tab.\,\ref{tab: exp_main}}  shows the effectiveness of our proposed {\ours} method vs. VP baselines (RLM-VP and FLM-VP) on diverse source models and target datasets. 
For comparison, we also present the model finetuning performance on target datasets using LP or FF. 
It is worth noting that  FLM-VP typically outperforms RLM-VP as the latter only uses a random label mapping to guide the learning of prompts \cite{tsai2020transfer}.
Thus, we only show the results of RLM-VP when using ResNet-18.

As shown in  \textbf{Tab.\,\ref{tab: exp_main}},  our proposed method ({\ours})   consistently outperforms other VP baselines by a large margin in nearly all the data-model setups, \textit{e.g.}, 7.9\%, 6.7\% and 6.5\% accuracy improvement over FLM-VP in the target dataset Flowers102, CIFAR100, GTSRB, respectively. \textit{In addition}, we note that model finetuning is typically more effective in transfer learning than prompting methods, consistent with existing work \cite{bahng2022visual}. This is not surprising as source models are allowed for modification, and the trainable parameter size increases (as evidenced by `Parameter Size' in \textbf{Tab.\,\ref{tab: exp_main}}). As will be evident in \textbf{Sec.\,\ref{sec: CLIP_LM}}, the accuracy of VP can be further improved if a language-vision source model is used.
Nonetheless, in the target dataset ABIDE, prompting methods can outperform the full model finetuning method (FF). Compared to other standard transfer learning tasks for image classification,  ABIDE was a newly-proposed medical dataset in \cite{tsai2020transfer}, which converts the original 1D numerical medical input sequences to image-alike data formats (\textit{i.e.}, brain-regional correlation graphs). The size of this dataset is extremely small due to the high cost of collecting data in the medical area, which restricts the performance of LP and FF. In contrast,   VP is uniquely suited for this setting.
\textit{Lastly}, in the model finetuning paradigm, a source model with larger capacity typically yields a better target task accuracy, \textit{e.g.}, the finetuning results of ResNet-50 vs. ResNet-18. However, this belief might not hold in the VP paradigm. As we can see, the prompting-induced target accuracy decreases under ResNet-50 in the target datasets Flowers102, EuroSAT, CIFAR100, and ABIDE.

Additionally, \textbf{Tab.\,\ref{tab: time_consumption}} in Appendix shows that ILM-VP takes a bit more run time than FLM-VP and LP, but is faster than FF. This is not surprising since the former adopts alternating optimization with a bit higher computation complexity than ordinary single-level minimization. 
Recently, the concurrent work \cite{wu2022unleashing} shows that properly re-sizing images before integrating with a VP could further boost the performance on a downstream task. We also find the same benefit of image re-sizing to VP on CIFAR10/100, GTSRB, and SVHN datasets (\textit{e.g.}, up-scaling the original image size to $128\times 128$) However, for ease of comparison with existing VP baselines ({RLM-VP} \cite{elsayed2018adversarial}), our experiments do not apply the image re-sizing trick to VP. 

\vspace{-5mm}
\paragraph{LM is key to improving the accuracy of VP.}
Next, we peer into the influence of LM on target prediction accuracy per class when using {\ours}.
In \textbf{Fig.\,\ref{fig: cls_acc}}, we demonstrate the testing accuracy improvements (over the FLM-VP baseline) of prompt-injected datapoints, belonging to 10 classes with the highest improvements selected from the target datasets Flowers102 and OxfordPets, respectively. Note that OxfordPets shares the most similar label space with ImageNet (\textit{e.g.} they both have beagles, boxers, bassets, etc.).
We use  $*$ in \textbf{Fig.\,\ref{fig: cls_acc}} to mark 
  target data classes whose source labels are remapped during {\ours}, and list non-$*$ marked target data classes whose source labels retain the same as FLM-VP. 
 We observe that target classes with large accuracy improvements typically require ILM. 
 This justifies the benefit of target-source label re-mapping during prompt learning. 
In addition, we note that the source labels of target classes (\textit{e.g.}, `yorkshire' in OxfordPets)   are not re-mapped, but {\ours} can still bring in accuracy improvements.
This implies that
LM has a coupling effect on all classes and 
the  BLO  framework \eqref{eq: BLO_ILM} enables us to improve LM as well as prompt learning in an interactive manner. 

Further, \textbf{Fig.\,\ref{fig: ILM_gradient}} shows the training dynamics of {\ours} vs. training epoch number 
and its convergence to the stable, high-explainable, and high-accurate visual prompt.
As we can see,  the mapped source label for a target class is updated at the early training epochs of {\ours}, but tends to converge at the later training phase. A similar trend holds for the convergence of LM difference between two adjacent epochs and the VP training loss. In addition, we can see that the VP pattern and the LM are updated jointly. 
Furthermore, the explainability of mapped source labels grows as the training proceeds. For example, the target label `Marigold' shares a similarity with the source label `Orange' in color and shape, as visualized by EBE-identified examples.
It is worth mentioning that EBE facilitates us to directly link the source dataset and the target dataset, and thus helps us to better understand the rationale behind VP. We refer readers to Appendix\,\ref{sec: explanation_by_example_analysis} for more EBE results.

\begin{figure}[t]
    \begin{minipage}[t]{0.48\columnwidth}
    \centering
    \includegraphics[width=\columnwidth]{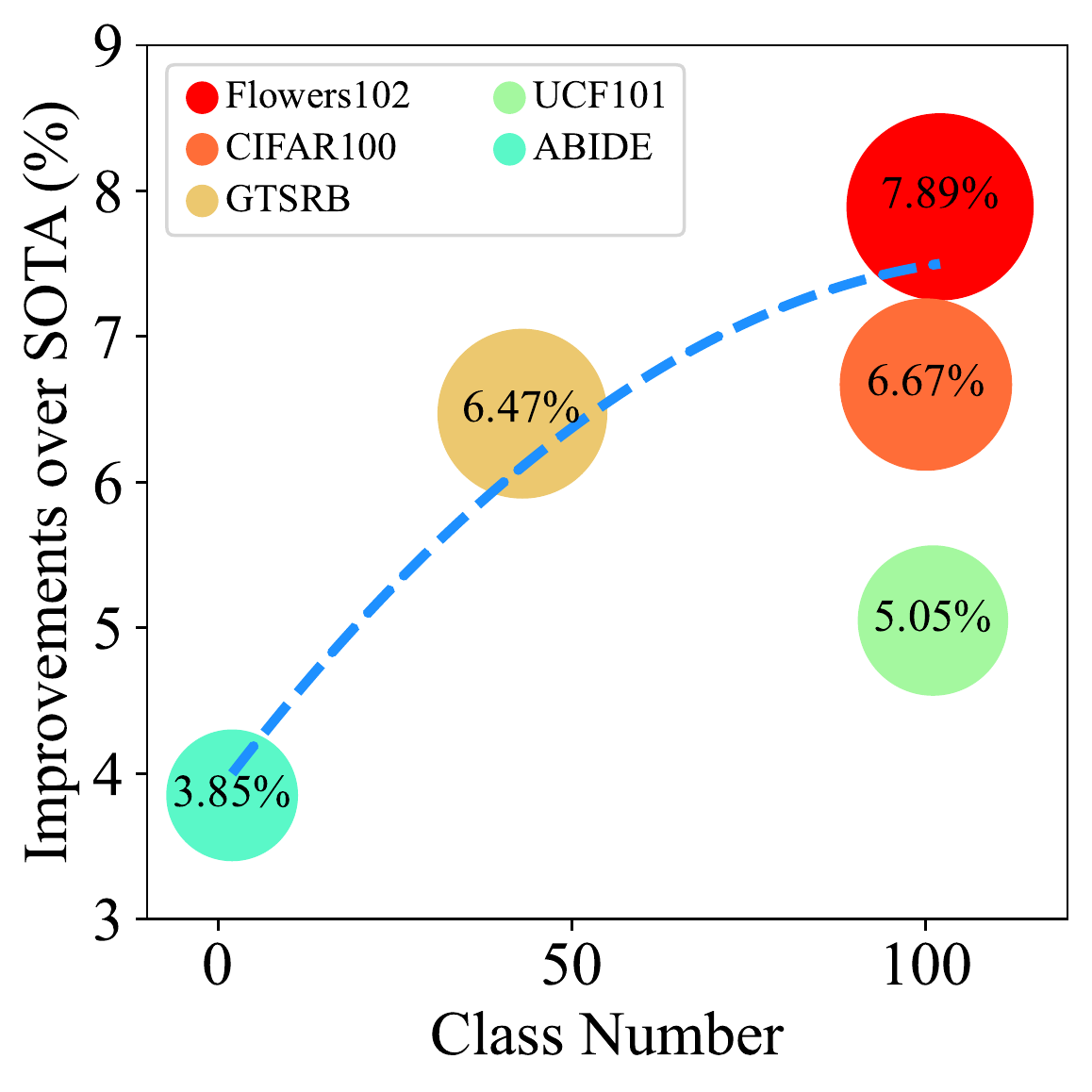}
    \caption{\footnotesize{{\ours}'s improvements over FLM-VP on representative datasets (datasets with improvements over 3\%) using  ResNet-18. The dashed line is the fitted curve.}}
    \label{fig: dataset_capacity_scatter}
\end{minipage}\hfill
\begin{minipage}[t]{0.48\columnwidth}
    \centering
    \includegraphics[width=\columnwidth]{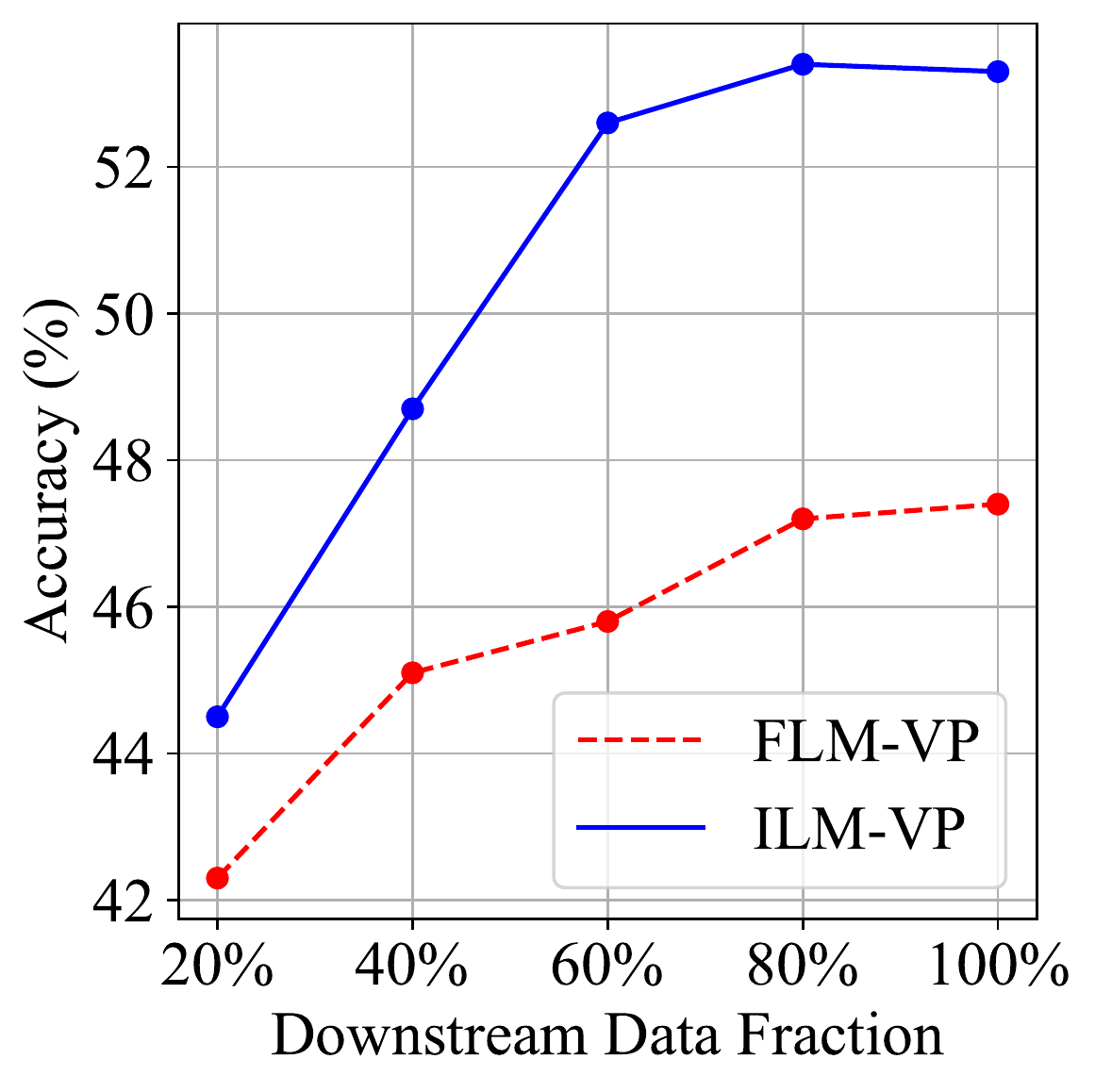}
    \caption{\footnotesize{{\ours} and FLM-VP performance on different fractions of GTSRB dataset (43 classes and more than 900 training samples per class) using ResNet-18. }}
    \label{fig: downstream_size}
    \end{minipage}\hfill
    \vspace{-8mm}
\end{figure}

\vspace{-5mm}
\paragraph{How target dataset scale affects VP?}
Through our experiments over a large number of target datasets, we find that {\ours} becomes more powerful when it comes to tasks with a larger target label space. For example, \textbf{Fig.\,\ref{fig: dataset_capacity_scatter}} shows the target datasets with at least 3\%  accuracy improvement using {\ours} compared with  FLM-VP on ResNet-18. As we can see, target datasets with the highest number of target classes correspond to the most significant accuracy improvement brought by {\ours}. 
Next,  we fix the target dataset and study how VP behaves at different downstream training dataset sizes. Here we choose GTSRB as the target task since GTSRB contains a sufficient amount of training data and thus facilitates us to conduct training dataset partition.   \textbf{Fig.\,\ref{fig: downstream_size}} compares the performance of {\ours} with FLM-VP vs. target training dataset size (training from 20\% to 100\% of the entire set). As we can see, {\ours} consistently outperforms the baseline FLM-VP and the improvement becomes more significant as the data scale grows. 

\vspace{-4mm}
\section{Extension: LM in Text Domain for  CLIP}
\label{sec: CLIP_LM}
\vspace{-2mm}

In the previous sections, we show that LM could play a vital role in VP when reprogramming a pre-trained vision model to conduct downstream targeted vision tasks.  In this section, we shift our focus from the vision source model to the vision-language model, specific to CLIP (contrastive language–image pretraining), which has received increasing attention in the area of VP \cite{bahng2022visual}. We will show that although CLIP does \textit{not} require source-target {mapping} across \textit{image labels} (due to its multi-modal learning architecture), the proposed idea on iterative LM can be extended to conduct \textit{text prompt selection} to improve target task accuracy.  

\vspace{-5mm}
\paragraph{LM for CLIP.} Different from the vision-only model,   CLIP can directly take target data labels as its textual inputs so as to mitigate the issue of source-target label mapping; see %\textbf{Fig.\,\ref{fig: CLIP_LM}}  and \SL{[remove this figure?]} 
\textbf{Fig.\,\ref{fig: CLIP_compare}} for an illustration. In this setting, LM seems redundant. However, this is only applied to   VP in the image domain. We argue that CLIP still needs implicit LM for \textit{text labels}, considering the diversity of text prompts (\textbf{TPs}) \cite{radford2021learning}. That is, CLIP can incorporate a text label into different context prompt templates (81 templates) suggested in \cite{radford2021learning} to create multiple text label instances. For example, the target label `dog' can be combined with context prompts `A photo of a big \{label\}' and `A photo of a small \{label\}'. Thus, given $m$ context prompts and $\Ktgt$ target data labels, we can create $m \Ktgt$ `virtual source labels', which should be mapped to $\Ktgt$ target labels. Thus, the LM problem arises, and its optimal solution characterizes the optimal prompt selection in the text domain for prompted image data.  Similar to the BLO method \eqref{eq: BLO_ILM}, we can bake iterative LM into VP using the CLIP model by replacing the lower-level image label mapping with the context-fused text label mapping. BLO then gives a unified prompt learning framework that can be easily compatible with CLIP. 
We refer readers to Appendix\,\ref{sec: clip} for more implementation details.

\vspace{-6mm}
\paragraph{LM improves VP's accuracy using CLIP.}
In \textbf{Tab.\,\ref{tab: exp_clip}}, we demonstrate the performance of the VP-driven CLIP model on several challenging target tasks shown in \textbf{Tab.\,\ref{tab: exp_main}}, \textit{e.g.}, Flowers102 and DTD.
%In Tab.\,\ref{tab: exp_clip}, we demonstrate that 
We term our method `VP+TP+LM', where the BLO-enabled LM method is called to map `virtual source labels' (\textit{i.e.}, the combination of context prompt template and target label) to realistic target image labels.
For comparison, we also present the performance of 
the baseline method termed `VP + TP' \cite{bahng2022visual}, which uses a pre-defined, fixed context prompt template `This is a photo of a \{label\}' when generating a visual prompt for CLIP. 
As we can see, 
our proposal consistently outperforms the baseline by a substantial margin. For example,  we obtain $13.7\%$ and $7.1\%$ accuracy gain in Flowers102 and DTD respectively.
In addition, we find that LM  brings in the interpretability merit: Our selected context prompt templates have better semantic meaning than the one used by the baseline. For example, VP for Flowers102 selects the text prompt `a close-up photo of a \{\}' instead of `This is a photo of \{\}' for the target image with the label `buttercup'. Another example is that VP for CIFAR10  prefers the text prompt `a pixelated photo of a \{\}'. In particular, we observe that in domain shift datasets (ImageNet-R and ImageNet-Sketch), the selected prompts can exhibit the domain information. More explainable results can be found in \textbf{Fig.\,\ref{fig: CLIP_more_results}}.

\begin{table}[htb]
\vspace{-3mm}
\centering
%\vspace*{-2em}

\resizebox{\columnwidth}{!}{%
\begin{tabular}{c|c|cc}
\toprule[1pt]
\midrule
\multirow{2}{*}{Methods} &
\multicolumn{1}{c|}{VP+TP} &
\multicolumn{2}{c}{Ours (VP+TP+LM)}   \\
& \multicolumn{1}{c|}{Acc(\%)}  & \multicolumn{1}{c}{Acc(\%)} & \multicolumn{1}{c}{
Examples of context prompt template $\to$ target label
}  \\
\midrule
Flowers102
& 70.0
& \textbf{83.7}
& a close-up photo of a \{\} $\to$ buttercup
\\
DTD
& 56.8
& \textbf{63.9}
& graffiti of a \{\} $\to$ blotchy
\\
UCF101
& 66.0
& \textbf{70.6}
& a \{\} in a video game $\to$ baseball pitch
\\
Food101
& 78.9
& \textbf{79.1}
& a photo of the dirty \{\} $\to$ crab cake
\\
SVHN
& 89.9
& \textbf{91.2}
& a photo of a \{\} $\to$ 7
\\
EuroSAT
& 96.4
& \textbf{96.9}
& a pixelated photo of a \{\} $\to$ river
\\
StanfordCars
& 57.2
& \textbf{57.6}
& the toy \{\} $\to$ 2011 audi s6 sedan
\\
SUN397
& 60.5
& \textbf{61.2}
& a photo of a large \{\} $\to$ archive
\\
CIFAR10
& 93.9
& \textbf{94.4}
& a pixelated photo of a \{\} $\to$ ship
\\
ImageNet-R
& 67.5
& \textbf{68.6}
& a rendition of a \{\} $\to$ gold fish
\\
ImageNet-Sketch
& 38.5
& \textbf{39.7}
& a sketch of a \{\} $\to$ eagle
\\
\midrule
\bottomrule[1pt]
\end{tabular}%
}
\caption{\footnotesize{Results of our CLIP-based prompt learning `VP+TP+LM' and the baseline `VP+TP' \cite{bahng2022visual} (restricted to using text prompt template ``This is a photo of a \{\}") over 11 target datasets. In each cell, the target task accuracy (\%) is shown along with examples of LM in the text domain. Our method with higher accuracy than SOTA is marked in \textbf{bold}.
}}
\label{tab: exp_clip}
\vspace*{-2.5mm}
\end{table}
\vspace{-5mm}
\section{Conclusion}
\label{sec: conclusion}
\vspace{-3mm}
This paper unveils LM's importance in the VP framework. Inspired by the prediction dynamics in optimizing VP, we formalize the VP problem through the lens of BLO (bi-level optimization). Upon our formalization, we propose a novel {\ours} algorithm to jointly optimize the input pattern training and the LM function. Across 13 datasets, we show our method's significant accuracy improvement over the SOTA VP baselines with graceful interpretability. Further, we extend our method to CLIP to improve its downstream task performance. % We believe our results provide useful insights into the optimal design of VP.
\vspace{-3mm}
\section{Acknowledgement}
\vspace{-3mm}
The work of A. Chen, Y. Yao, Y. Zhang, and S. Liu was supported by the DARPA RED program and National Science Foundation (NSF) Grant IIS-2207052.

%\clearpage
%%%%%%%%% REFERENCES
{\small
\bibliographystyle{unsrt}
\bibliography{ref.bib}
}

\clearpage
\appendix
\onecolumn
\setcounter{section}{0}

\section*{Appendix}

\setcounter{section}{0}
\setcounter{figure}{0}
\makeatletter 
\renewcommand{\thefigure}{A\arabic{figure}}% Figure counter representation
\renewcommand{\theHfigure}{A\arabic{figure}}% Hyperref figure hyperlink hook
\renewcommand{\thetable}{A\arabic{table}}
\renewcommand{\theHtable}{A\arabic{table}}

\makeatother
\setcounter{table}{0}

\section{VP Preliminary}
\label{sec: vp_preliminary}

\begin{figure}[htb]
    \centering
    \includegraphics[width=0.6\columnwidth]{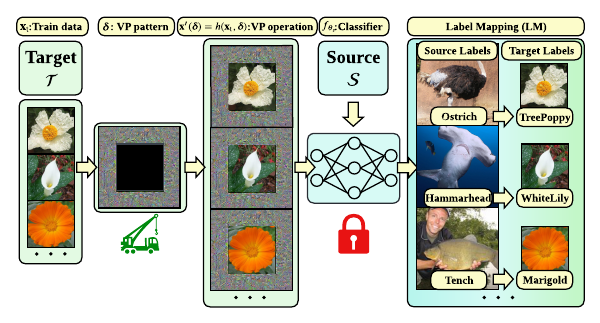}
    \caption{\footnotesize{{Overview of a vanilla VP pipeline. VP consists of (1) input prompting operation to incorporate prompt pattern into target data; (2) prompt generation by optimizing prompt pattern with the frozen source classifier; (3) label mapping for the upstream classifier to execute the downstream task.}}}
    \label{fig: VP_overview}
\end{figure}

Fig.\,\ref{fig: VP_overview} shows the pipeline of the vanilla VP method. The method has three main procedures: (1) input prompting operation to inject each image of the target dataset into the VP pattern; (2) prompt generation to optimize the VP pattern with the fixed, pre-trained source classifier; (3) pre-defined label mapping from the source data labels to the target data labels.

\section{Algorithm Overview of Our Proposal: {\ours}}

\begin{figure}[htb]
    \centering
    \includegraphics[width=0.6\columnwidth]{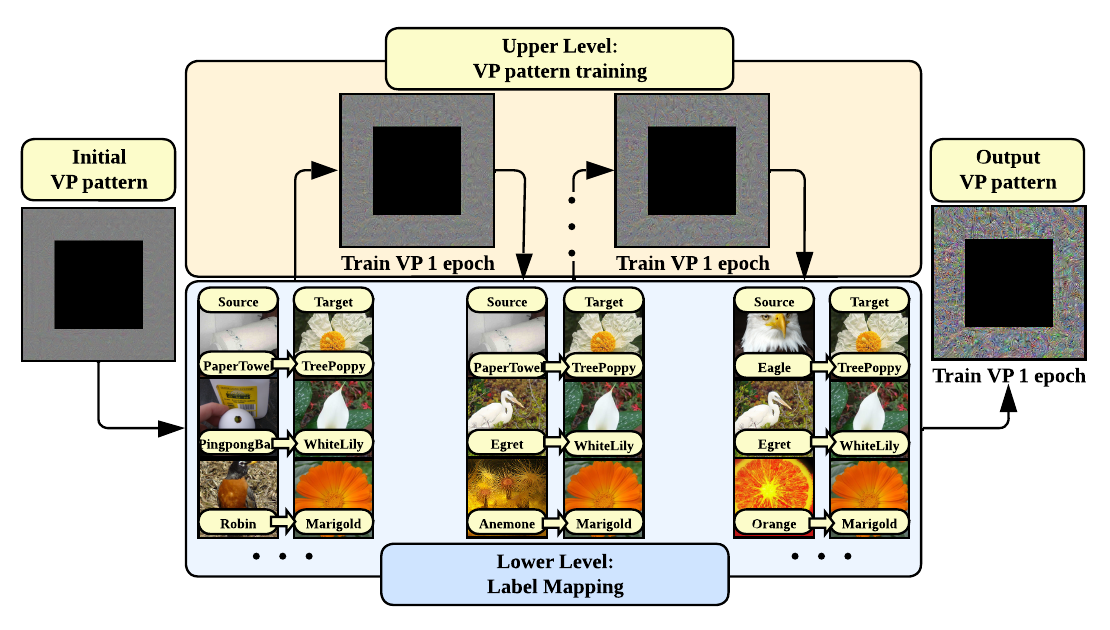}
    \caption{\footnotesize{Algorithm overview of {\ours}. Alternating optimization is iteratively  executed between the upper-level prompt generation and the lower-level LM. This BLO process can progressively improve both downstream task accuracy and LM interpretability.}}
    \label{fig: algorithm_overview}
\end{figure}

Fig.\,\ref{fig: algorithm_overview} shows the bi-level optimization pipeline of our proposal {\ours}. The method has two levels. (1) Lower level: Given the VP pattern from the previous epoch, use FLM technique to update the label mapping for each target class. (2) Upper level: Given the label mapping, call SGD to update the  VP pattern.

\clearpage
    
\section{Explanation by Examples}

\begin{figure}[htb]
    \centering
    \includegraphics[width=\columnwidth]{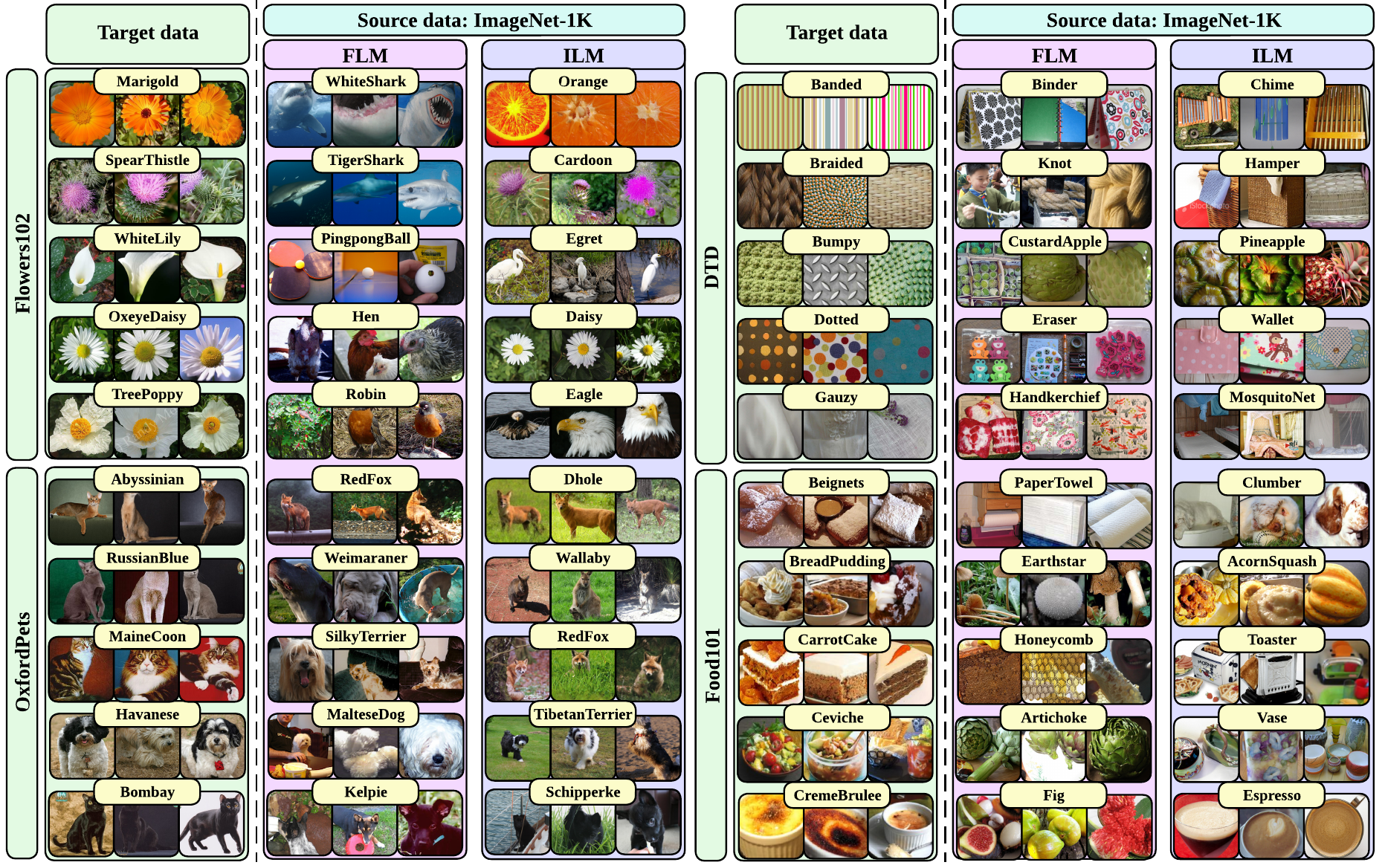}
    \caption{\footnotesize{LM method comparison: FLM vs ILM, by visualization of the target dataset (1) FLowers102, (2) OxfordPets, (3) DTD, and (4) Food101 with the source dataset ImageNet-1K using pretrained ResNet-18. ILM consistently finds more interpretable LM than FLM, in terms of colors, scenes, shapes, and textures. The four datasets are chosen due to the resolution for visualization and the accuracy improvement by our method compared to the prior art.}}
    \label{fig: more_FLM_ILM}
\end{figure}

\begin{figure}[h]
\centering
 \includegraphics[width=0.95\textwidth]{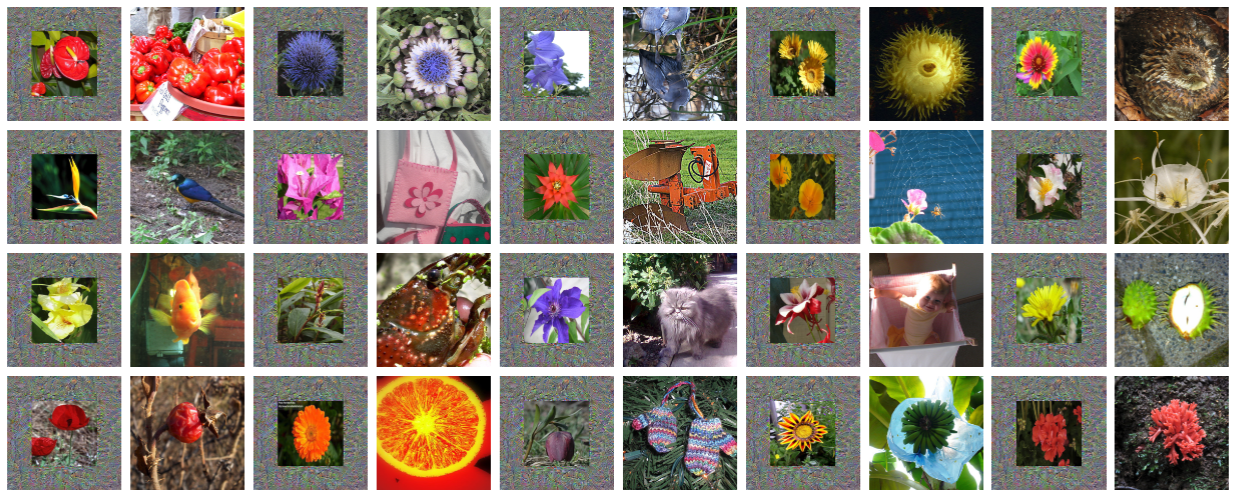}
  \caption{\footnotesize{More EBE examples in Flowers102. Every two pictures refer to one target class image and one corresponding source class image.}}
  \label{fig: ebe_rebuttal}
\end{figure}

\label{sec: explanation_by_example_analysis}
\paragraph{Implementation details.} Technically, given a test sample, EBE (explanation by example) finds the image from the training set which has the highest cosine similarity with the test datapoint. The cosine similarity is evaluated in the feature space, which is typically chosen as the activations of the penultimate layer of the neural network. We use a target task image integrated with the learned VP as the input sample and search for the most similar  training ImageNet samples in the mapped source class. The top-3 most similar  source examples are used    to explain the prediction using the pre-trained source model over the VP-injected test sample.

\paragraph{Discussion.} In VP, the source network preserves the full information of the source training dataset. Thus, EBE in VP makes a bridge between   the source dataset and the target dataset, which enables us to better understand the prediction ability of   VP.  Fig.\,\ref{fig: more_FLM_ILM} shows the EBE results of FLM-VP and {\ours}. Here we can   see that   {\ours} discovers more semantically-similar target-source pairs.
For example, `Oxeye Daisy' from Flowers102 is directly correlated with `Daisy' using ILM rather than `Hen' using FLM; `Bombay' from OxfordPets is correlated with a black dog feed `Schipperke' rather than a brown one `Kelpie'; `Banded' from DTD shares similar string with `Chime' using ILM rather than `Binder' selected by FLM; `Bread Pudding' from Food101 is mapped from `Acorn Squash' with similar textures and food property using ILM instead of `Earthstar' using FLM.
We argue that this high interpretability enabled by our proposed {\ours} method contributes to the analysis of transfer learning. 
Recent work on transfer learning has found that the source dataset has class-wise positive or negative influence in downstream tasks\cite{jain2022data}. Not surprisingly, they found a positive source class that shares some similar semantic features with one class in the target set. This is also revealed by our method (\textit{e.g.}, `Orange' mapped to `Marigold'). Since the method used in \cite{jain2022data} is computationally heavy, our method could be a light alternative to identify the most beneficial source class for each downstream class. 
We show more EBE results in Fig.\,\ref{fig: ebe_rebuttal}.

\clearpage

\section{VP Using Vision Models: Additional Experiments and Details}
\label{sec: additional_experiments_and_details}
\paragraph{Datasets.}
We focus on downstream image classification tasks in the target domain. To highlight a few of them, OxfordPets shares many same labels as ImageNet (\textit{e.g.}, beagle, boxer) while some datasets like GTSRB contain image classes quite different from ImageNet. CIFAR10 possesses a large training set (50000 images) yet a small label space (10 classes) while SUN397 has a rather small training set (15888 images) yet a huge label space (397 classes). In addition,  image classes in StanfordCars share quite similar features (\textit{e.g.}, different models and years of cars) while DTD contains rather different data features. Further, ABIDE is a medical dataset that converts the original 1D numerical input sequences to image-alike data formats.
We show each dataset's attributes in Tab.\,\ref{tab: dataset}.

\paragraph{Implementation details.} (Code link is available in the zipped supplementary file.)
We have RLM-VP (random label mapping based visual prompt), FLM-VP (frequency label mapping based visual prompt), LP (linear probing) and FF (full finetuning) as our baselines. We follow the same implementation in \cite{tsai2020transfer} to achieve the  frequency-based label mapping (FLM). In {\ours}, we execute   FLM   at the beginning of each epoch. To train VP, we choose Adam optimizer and use a learning rate of 0.01 determined by a multi-step decaying scheduler. The total number of trianing epochs is 200. 
%For datasets like CIFAR10 and GTSRB which have fixed image resolution, we use the original resolution of the target task images.
For datasets containing images of different resolutions (\textit{e.g.,} UCF101, Flowers102), we resize target task images into a fixed resolution. Target task resolutions and batch sizes of all datasets are shown in Tab.\,\ref{tab: dataset}. For LP and FF, we resize the target task images into 224$\times$224. In LP, we use the Adam optimizer, a multi-step decaying scheduler and a learning rate of 0.1. In FF, we use the Adam optimizer, a cosine-annealing scheduler and a learning rate of 0.01.  In the FF experiment, we   add data augmentation (\textit{i.e.,} Random Crop, Random Flip) and weight decay (with the magnitude of $5\times10^{-4}$).

\begin{table}[htb]
\vspace{-2mm}
\centering
%\vspace*{-2em}

\resizebox{0.7\columnwidth}{!}{%
\begin{tabular}{c|c|c|c|c|c}
\toprule[1pt]
\midrule
Dataset & Train Size & Test Size & Class Number & Batch Size & Rescaled Resolution \\
\midrule
Flowers102
& 4093
& 2463
& 102
& 256
& 128$\times$128
\\
DTD
& 2820
& 1692
& 47
& 64
& 128$\times$128
\\
UCF101
& 7639
& 3783
& 101
& 256
& 128$\times$128
\\
Food101
& 50500
& 30300
& 101
& 256
& 128$\times$128
\\
SVHN
& 73257
& 26032
& 10
& 256
& 32$\times$32
\\
GTSRB
& 39209
& 12630
& 43
& 256
& 32$\times$32
\\
EuroSAT
& 13500
& 8100
& 10
& 256
& 128$\times$128
\\
OxfordPets
& 2944
& 3669
& 37
& 64
& 128$\times$128
\\
StanfordCars
& 6509
& 8041
& 196
& 256
& 128$\times$128
\\
SUN397
& 15888
& 19850
& 397
& 256
& 128$\times$128
\\
CIFAR10
& 50000
& 10000
& 10
& 256
& 32$\times$32
\\
CIFAR100
& 50000
& 10000
& 100
& 256
& 32$\times$32
\\
ABIDE
& 931
& 104
& 2
& 64
& 200$\times$200
\\
\midrule
\bottomrule[1pt]
\end{tabular}%
}
\caption{\footnotesize{Dataset attributes and training configs through 13 target image-classification datasets.}}
\label{tab: dataset}
\end{table}

\begin{table}[htb]
\centering

\resizebox{0.5\columnwidth}{!}{%
\begin{tabular}{c|c|ccc}
\toprule[1pt]
\midrule
Architecture
& ILM-VP
& FLM-VP
& LP
& FF
\\
\midrule
ResNet18
& 20 
& 14 
& 15 
& 33 
\\
ResNet50
& 26 
& 17 
& 17 
& 49 
\\
ResNeXt-101-32x8d
& 62 
& 45 
& 47 
& 104 
\\
\midrule
\bottomrule[1pt]
\end{tabular}%
}
\caption{\footnotesize{V100 run-time on Flowers102 in minutes.}}
\label{tab: time_consumption}
\end{table}

\clearpage
\section{VP Using CLIP: Additional Experiment Details and Results}
\label{sec: clip}

\begin{figure}[htb]
    \centering
    \includegraphics[width=0.6\columnwidth]{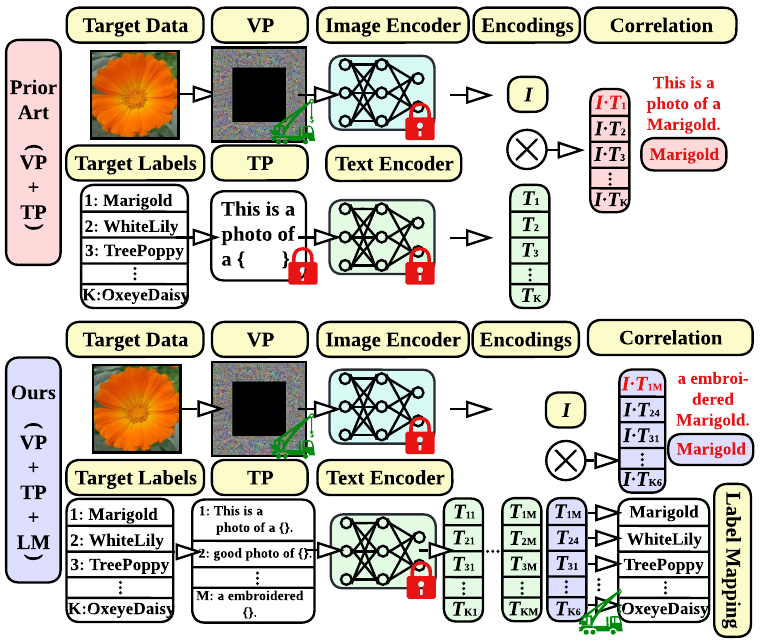}
    \caption{\footnotesize{Overview of CLIP-based prompt learning: our proposal vs. SOTA method.  Our proposal incorporates label mapping into the text prompt selection of CLIP.}}
    \label{fig: CLIP_compare}
\end{figure}

\paragraph{Implementation details.}
Fig.\,\ref{fig: CLIP_compare} shows the difference between the prior art (upper one) and our proposal (lower one). In the prior art, `This is a photo of a \{\}' is the only context prompt selected for all different target labels, then correlated with the image encoding to accomplish the target image classification task. In contrast, our proposal considers all  81 different text prompts introduced in the original CLIP setting. We incorporates the  label mapping technique to map virtual source labels (given by the combinations of context prompts and target classes) to target labels. We strictly follow the implementation detailed in \cite{bahng2022visual} for the VP+TP baseline. The only difference is that we train 200 epochs instead of 1000 epochs. We observed that the performance is similar and 200 epochs' training is much more efficient. For our VP+TP+LM method, we first create multiple virtual source labels and then do FLM at the beginning of each epoch. The rest setting is  the same as the baseline.

\begin{figure}[htb]
    \centering
    \includegraphics[width=1\columnwidth]{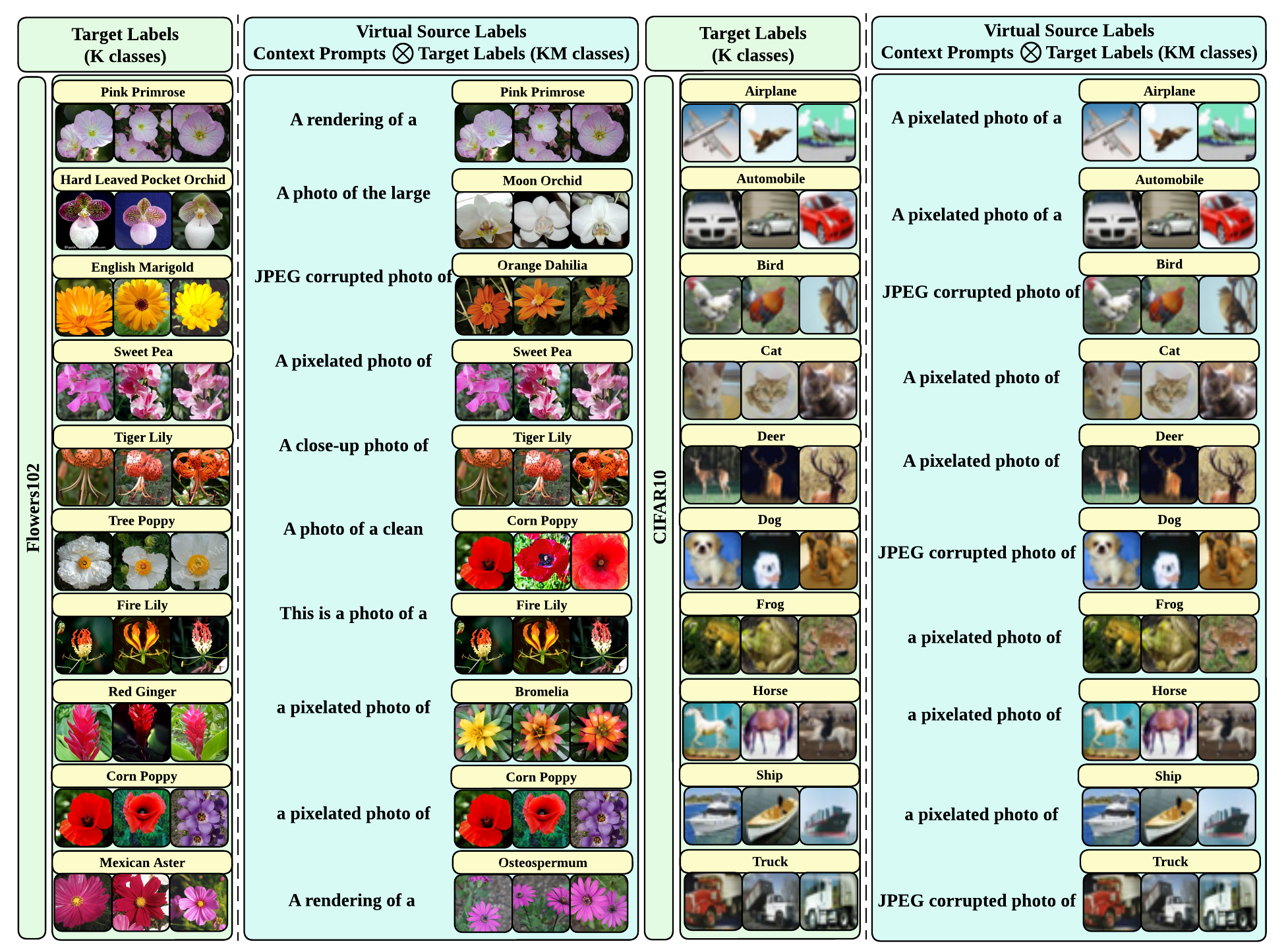}
    \caption{\footnotesize{Label mapping results of our proposed VP method for   CLIP. The presented two datasets 
    `Flowers102' and `CIFAR10' shows significant interpretability when using label mapping.}}
    \label{fig: CLIP_more_results}
\end{figure}

\paragraph{Interpretability merit of proposed VP+TP+LM method.}
We show in Fig.\,\ref{fig: CLIP_more_results} the interpretability of prompts found by our method. For example, in Flowers102, `Tiger Lily' chooses `A close-up photo of Tiger Lily' as its text prompt. This is reasonable as the dataset consists most of close-up photos of flowers.
%provides more camera information about images of `Tiger Lily' than `This is a photo of Tiger Lily' because the dataset consists most of close-up photos of flowers. 
% In DTD, the `Blotchy' class chooses `A sculpture of a Blotchy' for the same reason. In addition, there   also exist cross-class mappings. 
In CIFAR10, the "Airplane" class chooses "A pixelated photo of Airplane" for the same reason. 
In addition, there   also exist cross-class mappings. 
For example, `Hard Leaved Pocket Orchid' corresponds to `A photo of the large Moon Orchid'. We can  see that both `Hard Leaved Pocket Orchid' and `Moon Orchid' belong to `Orchid' and the former one is larger in size. 

% In DTD, `Banded' corresponds to `A cartoon Stripe' with the similar rationale to the above example.

\iffalse
\begin{minipage}{.8\linewidth}
\begin{algorithm}[H]
\caption{The proposed {\ours} algorithm}
\begin{algorithmic}[1]
  \State \textbf{Initialize:} Given target training set $\mathcal{T}_\mathrm{tr}$, pre-trained model $f_{{\bthetasrc}}$, prompt pattern initialization $\boldsymbol{\delta}_0$, and upper-level learning rate $\lambda$ for SGD
  \For{Epoch $n=0,1,\ldots,$}
\State \textbf{Lower-level label mapping}: Given $\boldsymbol{\delta}_{n-1}$,  call LM 
  for each target class $\ytgt$ in $\mathcal{T}_\mathrm{tr}$
\State \textbf{Upper-level prompt learning}: Given LM, call SGD to update prompt
$\bdelta_n \xleftarrow{} \bdelta_{n-1}$
\EndFor
\end{algorithmic}
\end{algorithm}
\end{minipage}
\fi

\end{document}